\documentclass{article} 
\usepackage{iclr2025_conference,times}

\usepackage{amsmath,amsfonts,bm}

\def\eqref#1{equation~\ref{#1}}

\def\1{\bm{1}}

\DeclareMathAlphabet{\mathsfit}{\encodingdefault}{\sfdefault}{m}{sl}
\SetMathAlphabet{\mathsfit}{bold}{\encodingdefault}{\sfdefault}{bx}{n}

\usepackage{hyperref}
\usepackage{url}
\usepackage{latexsym}
\usepackage[T1]{fontenc}
\usepackage[utf8]{inputenc}
\usepackage{microtype}
\usepackage{booktabs}
\usepackage{multirow}
\usepackage{fancyhdr}
\usepackage{amsmath}
\usepackage{amsthm}
\usepackage{adjustbox}
\usepackage{CJKutf8}

\usepackage{enumitem}
\usepackage{listings}
\lstdefinelanguage{prompt}{
    basicstyle=\small\ttfamily,
    commentstyle=\color{cyan},
    morecomment=[l]{//},
    moredelim=[is][\color{brown}\bfseries]{<<<}{>>>},
    moredelim=[is][\color{magenta}\bfseries]{[[[}{]]]},
    moredelim=[is][\color{orange}\bfseries]{===}{===},
    moredelim=[is][\color{blue}\bfseries]{|||}{|||},
    moredelim=[is][\bfseries]{**}{**},
}
\lstdefinelanguage{ioexample}{
    frame=shadowbox,
    rulesepcolor=\color{gray},
    framerule=0.5mm,
    rulesep=2mm,
    basicstyle=\small\normalfont,
    commentstyle=\color{cyan},
    morecomment=[l]{//},
    moredelim=[is][\color{green}\bfseries]{<<<}{>>>},
    moredelim=[is][\color{magenta}\bfseries]{[[[}{]]]},
    moredelim=[is][\color{orange}\bfseries]{===}{===},
    moredelim=[is][\color{olive}\bfseries]{|||}{|||},
    moredelim=[is][\bf]{:::}{:::},
    moredelim=[is][\it]{---}{---},
    moredelim=[is][\tt]{+++}{+++},
}

\newcommand{\datasetname}{\texttt{PRELUDE} }
\newcommand{\datasetnamens}{\texttt{PRELUDE}}

\usepackage{caption}  
\usepackage{wrapfig}

\renewcommand{\theexample}{\Roman{example}}

\newenvironment{exampletable}[2][]{
  \refstepcounter{example} 
  \begin{center}
    \noindent\textbf{\small Example~\theexample.}~#2 
  \ifx\relax#1\relax
  \else
    \label{#1}
  \fi
  
}{
  \end{center}
}

\title{\datasetnamens: A Benchmark Designed to Require Global Comprehension and Reasoning over Long Contexts}

\iclrfinaltrue

\author
{Mo Yu~$^{1\thanks{Equal contribution. Correspondence to:  \texttt{moyumyu@tencent.com,ttchungac@connect.ust.hk}.}}$, Tsz Ting Chung~$^{2\footnotemark[1]}$, Chulun Zhou~$^{3\footnotemark[1]}$, Tong Li~$^{3\footnotemark[1]}$, Rui Lu~$^{3\footnotemark[1]}$, Jiangnan Li~$^{1\footnotemark[1]}$, Liyan Xu~$^{1\footnotemark[1]}$, \\
\bf Haoshu Lu~$^{4}$, Ning Zhang~$^{1}$, Jing Li~$^{4}$, Jie Zhou~$^{1}$
\vspace{0.5em} 
\\
Project Page: {\textcolor{magenta}{\hypersetup{urlcolor=magenta} \url{https://gorov.github.io/prelude}}}
\vspace{0.5em} \\
\normalfont{\small $^{1}$WeChat AI, Tencent}\quad\normalfont{\small $^{2}$HKUST}\quad
\normalfont{\small $^{3}$CUHK}\quad
\normalfont{\small $^{4}$NJIT}
\vspace{0.5em} \\
\textcolor{red}{\textsf{\Large\textbf{Spoiler alert: We show it is possible to \emph{Measure Fluid}}}} \\
\textcolor{red}{\textsf{\Large\textbf{\emph{Intelligence} in natural language space.}}} 
}

\begin{document}

\maketitle
\begin{figure}[h]
  \includegraphics[width=0.52\columnwidth]{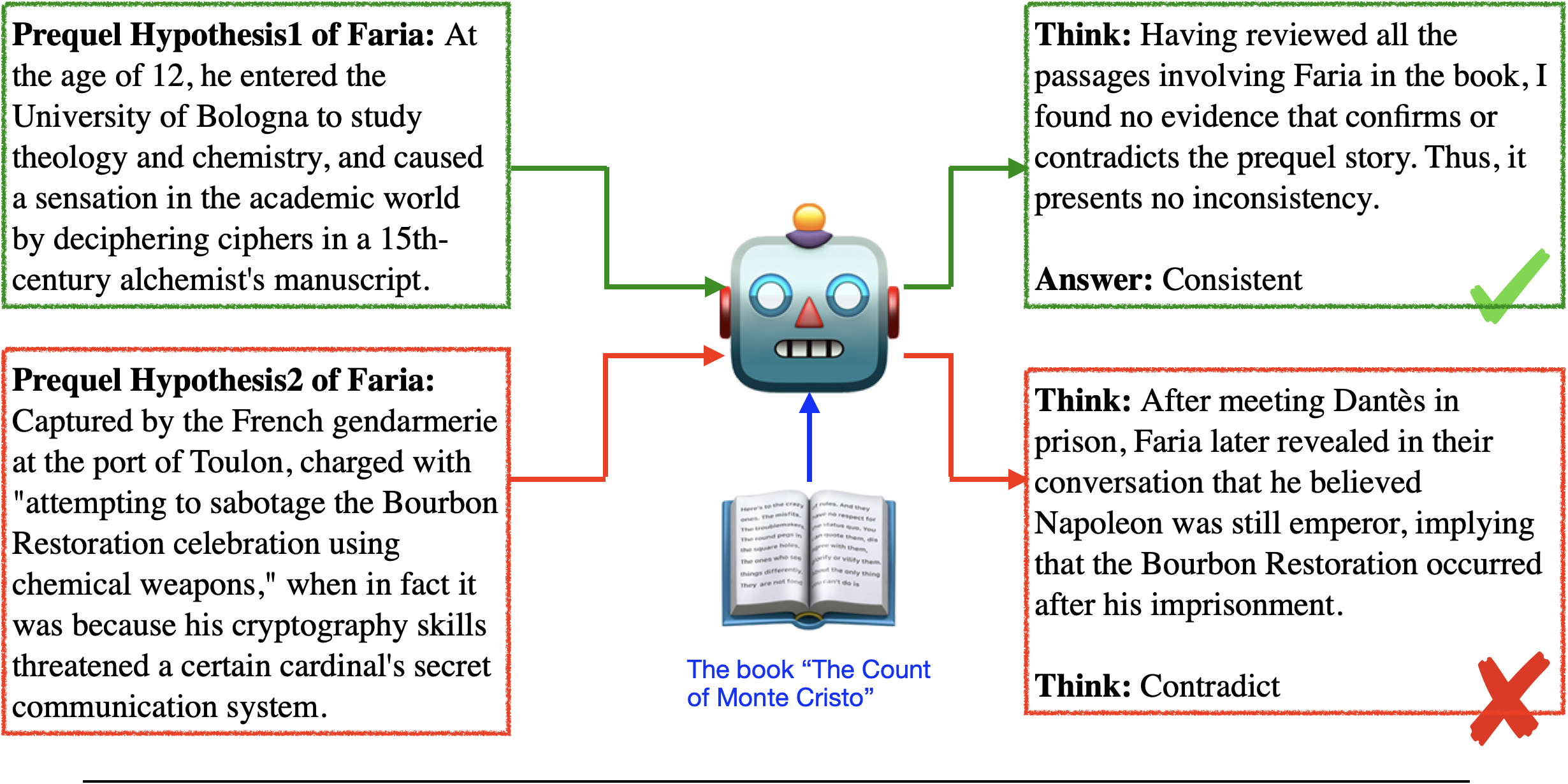}
  \hspace{0.1in}
  \includegraphics[width=0.45\columnwidth]{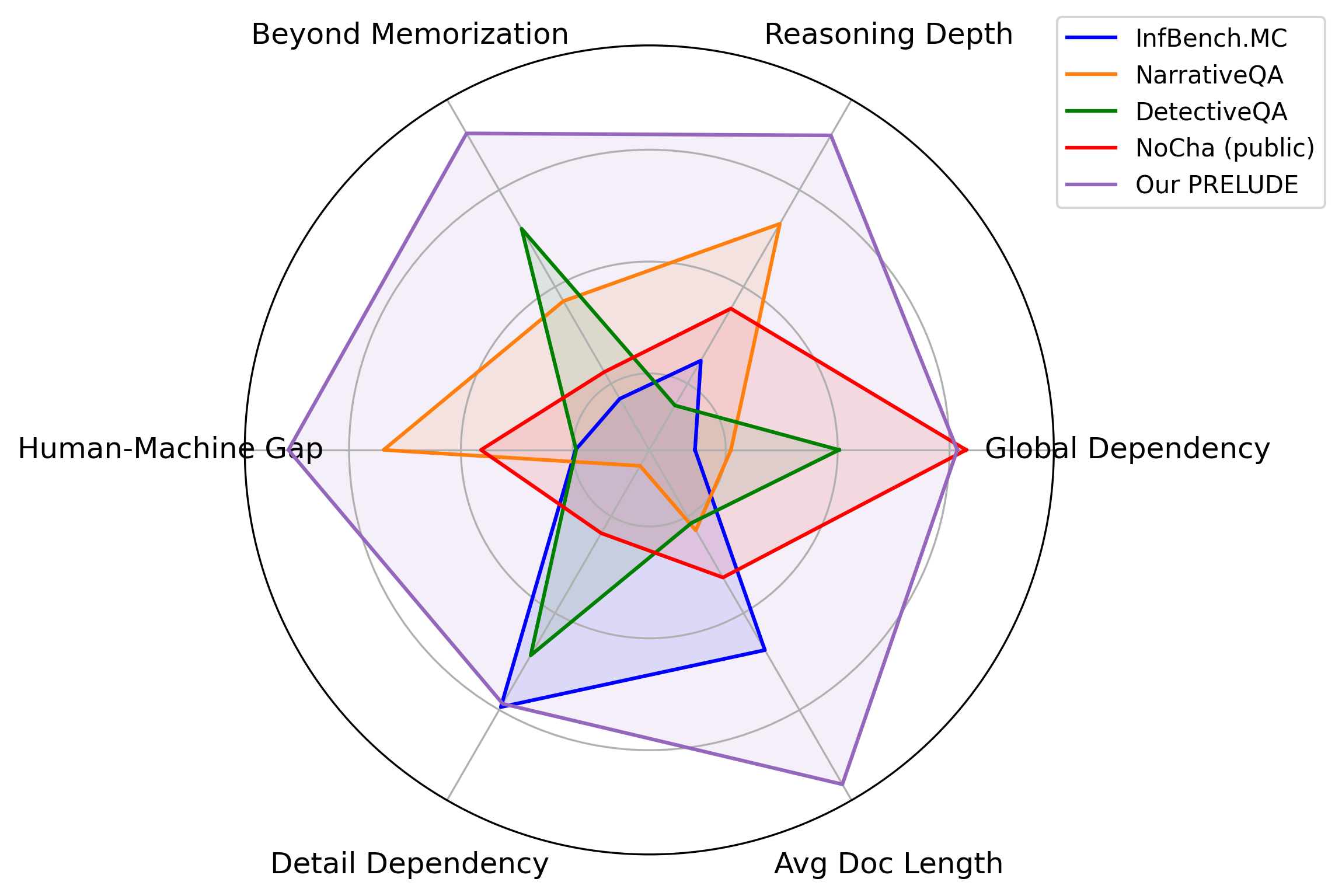}
  \caption{\small (Left) Two examples from our \datasetnamens. (Right) Comparison of existing benchmarks along the different criteria for long context understanding assessment. We report the normalized measure along each criterion, with the details presented in Appendix~\ref{app:criteria}.}
  \label{fig:intro}
\end{figure}

\begin{abstract}
We introduce \datasetnamens, a benchmark for evaluating long-context understanding through the task of determining whether a character's prequel story is consistent with the canonical narrative of the original book.
Our task poses a stronger demand for global comprehension and deep reasoning than existing benchmarks -- as the prequels are not part of the original story, assessing their plausibility typically requires searching and integrating information that is only indirectly related. Empirically, 88\% of instances require evidence from multiple parts of the narrative. 
Experimental results highlight the challenge of our task: in-context learning, RAG and in-domain training with state-of-the-art LLMs, and commercial {DeepResearch} services, lag behind humans by $>$15\%. 
A further human study reveals that models often produce correct answers with flawed reasoning, leading to an over 30\% gap in reasoning accuracy compared to humans.
These findings underscore the substantial room for improvement in long-context understanding and reasoning.
\end{abstract}


\section{Introduction}
\label{sec:intro}
The emergence of new LLM-driven applications, such as 
multi-document analysis~\citep{notebooklm,wang2024leave,hipporag}, personal assistants with chat histories~\citep{wu2024longmemeval,xu2025mem}, autonomous agents~\citep{hu2025owl,openai2025deepresearch,wang2025ragen}, and repository-level coding tools~\citep{jimenezswe24}, has created increasing demands for robust long-context understanding and reasoning.

To better support long inputs, many techniques have been proposed, primarily focusing on efficient attention mechanisms~\citep{xiong2021nystromformer,long-llama} and retrieval-augmented generation (RAG)~\citep{rag,xu2024narco,graphrag,hipporag,selfrag}.
Alongside these technical advances, there is a growing need for effectively evaluating long context understanding and reasoning capabilities.
To this end, several benchmarks have recently been introduced (see Section~\ref{sec:related}).
Building on this progress, recent work has extensively discussed the criteria that a strong benchmark for long-context understanding and reasoning should satisfy~\citep{press2023measuring, yu2023personality, liu2024lost, yen2024helmet, fang2024wrong, wu2025shifting}.
To rigorously evaluate a model’s capabilities in this domain, several key criteria have emerged as essential:
\begin{itemize}[noitemsep,nolistsep,leftmargin=*]
\item \textbf{Beyond Memorization.}
LLMs memorize content from pretraining, especially for popular texts~\citep{tirumala2022memorization,deletang2023language,ilya2023}, enabling answers without true comprehension. 
The existence of this shortcuts blurs the line between long-context understanding and mere activation of parametric knowledge memorized during pretraining.
As training data grows, this issue worsens. As a \emph{Necessity Condition}, a robust benchmark must prevent solutions based on memorization alone, ensuring full-context reasoning remains essential.

\item \textbf{Global Dependency.}
The task should require aggregating evidence scattered across the context or exhibiting global dependencies; otherwise, it reduces to a short-context problem focused on retrieval rather than true long-text understanding.

\item \textbf{Depth of Reasoning.}
Long-context reasoning should inherently require synthesizing multiple pieces of evidence and multi-step deduction. However, many existing benchmarks focus on shallow reasoning, such as decomposition or enumeration (\emph{e.g.}, counting facts~\citep{xu2025llm}, or sequential sub-questions~\citep{yang2018hotpotqa}), which reduces the need for global, multi-step inference and oversimplifies the task.

\item \textbf{Human-Machine Gap.}
To highlight essential capabilities that general-purpose intelligent systems should possess,
a benchmark should show a significant gap between humans and machines. Low human accuracy or agreement typically signals poor annotation quality or high subjectivity, undermining reliable evaluation.

\item \textbf{Beyond Summarization/Salience.}
Often overlooked but crucial, a strong benchmark should require attention to fine-grained details beyond high-level abstraction to remain challenging and meaningful.
Otherwise, it risks reducing to a summarization task that is solvable without long-context understanding~\citep{chang2023booookscore}. 

\end{itemize}

Figure~\ref{fig:intro} evaluates several widely-used benchmarks for story understanding across the aforementioned dimensions. 
It shows that all existing benchmarks fall short in at least one aspect, with a particularly notable limitation in mitigating the effects of \emph{Memorization} and encouraging \emph{Deep Reasoning}. Detailed evaluation metrics and experimental settings are provided in Appendix~\ref{app:criteria}.

To address these limitations, we propose a novel task format that enables comprehensive assessment across all the identified criteria. The task involves presenting the model with a list of hypothetical prequels for important supporting book characters, 
and asking it to determine whether each prequel is consistent with the canonical story.
Each hypothesis is presented as a concise bullet point (see Figure~\ref{fig:intro}) summarizing a proposed setting. These hypotheses are annotated by human experts who have read the canonical stories multiple times and performed literary analysis before. This annotation process is both efficient and high-quality, ultimately yielding a dataset of $\sim$1K labeled examples.
Intuitively, our task design naturally mitigates the key limitations observed in existing benchmarks:

First, the \emph{Memorization} shortcut is alleviated by construction, as the input prequels are newly generated and do not appear in the training data of any existing LLMs. The limited performance of OpenAI’s Deep Research further supports this, suggesting that it is difficult to locate human-summarized evidence on the internet to solve our task.

Second, our task encourages \emph{global reasoning}. This is because (1) determining whether a consistent prequel aligns with the canonical story typically requires aggregating evidence across the whole character story; and (2) contradictory prequels often involve inconsistencies that span several scattered events due to the narrative structure of the original work. Empirically, our annotation analysis reveals that 88\% of the examples in \datasetname require non-local evidence to resolve.

Finally, our task encourages \emph{deep reasoning}, because of the fact that the canonical story reflects non-immediate consequences of the prequels. To solve our task, LLMs must unfold the implications of a prequel and align them with the story, often requiring multi-step inference. For instance, the second example in Figure~\ref{fig:intro} involves reasoning that \emph{Faria was arrested when Napoleon was still emperor}, and then inferring a contradiction from the fact that \emph{the Bourbon Restoration removed Napoleon from power}. This kind of non-immediate causality resists shallow reasoning shortcuts that decomposes the problem into subquestions. 

We conduct extensive experiments on our proposed task using ICL, RAG, and DeepResearch across state-of-the-art commercial and open-source LLMs. The results reveal several key findings:
(1) The best-performing system lags human performance by over 15\%;
(2) LLMs often arrive at correct predictions with flawed reasoning, resulting in a reasoning accuracy gap of $>$30\% compared to humans;
(3) Our task cannot be solved by searching for information on the web, making the advanced DeepResearch systems struggle and underperform RAG;
(4) Supervised training and many-shot ICL yield no performance gains, highlighting LLMs’ intrinsic limitations in long-context reasoning.

These findings highlight that \datasetname requires deeper long-context reasoning capabilities beyond what current methods offer, pointing to important directions for future research.

\begin{table}[t]
\small
\centering
\resizebox{\columnwidth}{!}{
\begin{tabular}{l|p{4cm}|p{10cm}}
  \toprule
  \bf Type & \bf Definition & \bf Example \\
  \midrule
  Contradict -- Local & A detail in the original novel directly contradicts the prequel & \textbf{Book:} The Chronicles of Narnia\quad \textbf{Char:} Eustace Scrubb \newline \textbf{Prequel:} Born in London and raised as an only child, Eustace was taught to call his parents by their names —"Harold" and "Alberta".\newline \emph{The novel introduces Eustace as being born in Cambridge, England.} \\
  \midrule
  Contradict -- Global I & The prequel causes part of the original plot or the character’s motivations to become broadly unreasonable. & \textbf{Book:} The Count of Monte Cristo\quad \textbf{Char:} Faria \newline \textbf{Prequel:} He joined the Jesuits and was sent to Goa, Indian. During this time, he secretly studied the ancient Indian medical text "Charaka Samhita," which laid the groundwork for his later expertise in toxicology. 
  \newline
  \emph{(Throughout the story, Faria never demonstrates any expertise in medicine or toxicology. Moreover, he does not speak Hindi, making such an event unlikely.)}\\
  \midrule
  Contradict -- Global II & The prequel does not contradict the original plot but is inconsistent with the tone or stylistic setting of the novel & \textbf{Book:} Romance of the Three Kingdoms \quad \textbf{Char:} Xiahou Yuan \newline \textbf{Prequel:} He was taught magical speed-enhancing techniques by a mysterious hermit, which allowed him to march 1000 li in six days.\newline\emph{(The book is a historical novel with a largely realistic style, and it never depicts any general using magic or supernatural powers.)}\\
  \midrule
  Consistent -- Irrelevant & The prequel is not closely relevant to the original plot but doesn’t contradict it. It is a side story of the character. & \textbf{Book:} The Count of Monte Cristo\quad \textbf{Char:} Faria \newline \textbf{Prequel:} At the age of 12, he entered the University of Bologna to study theology and chemistry, and caused a sensation in the academic world by deciphering the ciphers in a 15th-century alchemist's manuscript.\newline\emph{(Faria’s academic background and influence are never depicted in the novel. However, this is consistent with his characterization as a learned man.)}\\
  \midrule
  Consistent -- Core & The prequel fills in missing detail without introducing contradictions. & \textbf{Book:} Harry Potter\quad \textbf{Char:} Sirius Black \newline \textbf{Prequel:} Rejecting pure-blood elitism, Sirius frequently showed admiration for Gryffindor and spent time with Muggles and so-called "blood traitors" to annoy his parents even before attending Hogwarts.\newline\emph{(Sirius is introduced as a key figure against pure-blood elitism, but little detail is provided. This fills in gaps without introducing contradictions.)} \\ 
  \bottomrule
\end{tabular}
}
\caption{\small{Definitions of our annotation labels in \datasetnamens.}}
\label{tab:definitions}
\end{table}

\section{Related Work}
\label{sec:related}

\paragraph{Tasks over Synthetic Long Contexts}

Following the needle-in-a-haystack task that examines LLM's in-context searching \citep{kamradt2023needle}, a line of works have since focused on probing the LLM ability to trace and utilize information pieces in stretched long context~\citep{hsieh2024ruler,li2025needlebenchllmsretrievalreasoning,yu2025sequential}, while others also fuse reasoning in their task design such as sorting or relation identification, beyond the mere retrieval aspect \citep{kuratov2024babilong,wang-etal-2024-ada,dong-etal-2024-bamboo,wang-etal-2024-leave,lee-etal-2025-ethic}. Besides these works that specifically stress-test long context utilization, other related LLM tasks could also reflect such ability, \emph{e.g.}, many-shot in-context learning \citep{agarwal2024manyshot,xu2024stresstesting,li2025longcontext}.

\paragraph{Realistic Long Context Understanding Tasks}
Orthogonal to those synthetic stress-testing tasks, another line of works target the more natural question answering settings for realistic long context evaluation, primarily utilizing stories in various domains, such as NarrativeQA~\citep{kovcisky2018narrativeqa}, NovelQA~\citep{DBLP:journals/corr/abs-2403-12766}, DetectiveQA~\citep{xu2025detectiveqa}, CharToM-QA~\citep{zhou2025essence}. Realistic long context QA has gained particular attention in many recent LLM benchmarks, such as LongBench~\citep{DBLP:conf/acl/BaiLZL0HDLZHDTL24,bai2024longbench}, XLBench~\citep{ni2024xl}, CLongEval~\citep{DBLP:conf/emnlp/QiuLHJZK24}, LooGLE~\citep{DBLP:conf/acl/LiWZZ24}, $\infty$ Bench~\citep{DBLP:conf/acl/ZhangCHXCH0TW0024}, LaRA~\citep{li2025lara}, \emph{etc}.

\paragraph{Document-Level Entailment}
Our task is closely related to fact verification over multiple documents or web sources, as exemplified by FEVER~\citep{thorne2018fever} and its extensions~\citep{wadden2020fact,DBLP:conf/acl/YinRX21,schlichtkrull2023averitec}.

Among this line of work, NoCha~\citep{DBLP:conf/emnlp/KarpinskaTLGI24} is the most relevant, as it also uses book narratives as context. However, a key distinction lies in the nature of the hypotheses:
NoCha uses summaries or conclusions of the original story, which often share semantic overlap with the canonical book. Therefore, this task design is vulnerable to memorization or summarization shortcuts, as shown in Figure~\ref{fig:intro}.
To mitigate this, NoCha uses recently published books. Yet as training corpora expand, newer LLMs inevitably become familiar with these texts, reducing the task's effectiveness. 
As shown in our experiments, while the public subset of NoCha has been largely conquered by LLMs, our subset comprising works from the same period or earlier remains challenging. This shows that our task is not becoming easier for models over time.

\section{The Proposed \datasetname Task}
\label{sec:def_ending}
In this section, we introduce the construction process of our \datasetname (PRequel Entailment for Long context Understanding and DEduction) task.

\subsection{Task Formulation}
Our task is formulated as binary classification.
The input consists of a book $\mathcal{B}$ that is split into $M$ consecutive chunks ${\mathcal{B}}=\{b_1, b_2, ..., b_{M}\}$; and a prequel $p$ for a character $c$, which is a short text describing an experience of the character prior to the story of $\mathcal{B}$ happens.
The task is then predict whether $p$ aligns with $\mathcal{B}$.
The labels to predict thus are $\{consistent, contradict\}$.

\subsection{Why Prequels?}

Our prequel entailment task is naturally a long-context understanding task and a form of everyday research task.
To solve the task, a model needs to judge whether a character’s prequel remains consistent with the behaviors and experiences throughout the narrative and makes counterfactual reasoning when necessary.
These requirements make our task well-suited for benchmarking long-context reasoning for the following desirable properties:

\begin{itemize}[noitemsep,nolistsep,leftmargin=*]
    \item \textbf{Natural long-context reasoning:} The task requires holistic understanding of a narrative arc, including tracking a character’s psychological continuity, goals, and situational influences across temporally distant events. 
    \item \textbf{Cognitive research practices in daily life:}
    While formal research is often confined to scientific domains, its core cognitive components, such as gathering evidence, forming hypotheses, and drawing conclusions, are deeply embedded in daily reasoning. 
    Our task scenario mirrors this real-life cognition, as humans frequently make similar judgments while watching films, reading novels, or engaging in social interactions.
    \item \textbf{Light dependency on background knowledge:} This task requires little reliance on external or specialized knowledge. 
    A reader with a full understanding of the story, even as a middle school student, can often make accurate judgments. As a result, the task emphasizes fluid intelligence rather than crystallized knowledge acquired through prior learning.
\end{itemize}

\begin{table*}[t!]
\small
\centering
\resizebox{\columnwidth}{!}{
\begin{tabular}{lllll||cc} 
\toprule
\bf Name& \bf Author & \bf Genre & \bf Lang. & \bf Public & \bf \#Char & \bf \#Instances \\
\midrule
The Count of Monte Cristo & Alexandre Dumas & Adventure fiction  & English & Yes & 2 & 54 \\
Demi-Gods and Semi-Devils & Louis Cha & Martial art fiction & Chinese & No&  3 & 78 \\
The Return of the Condor Heroes & Louis Cha & Martial art fiction & Chinese & No& 4 & {44}\\
Investiture of the Gods & Xu Zhonglin& Mythology & Chinese & Yes& 4 & 94\\
Romance of the Three Kingdoms & Luo Guanzhong & Historical fiction & Chinese & Yes& 2 & 28\\
Love in the Time of Cholera & Gabriel García Márquez & Magical realism & English & No& 1 & {15}\\
Pinball, 1973  & Haruki Murakami & Surrealism & Chinese & No& 2 & {22} \\
Rebecca & Daphne du Maurier & Gothic fiction& English & No& 3 & 82\\
In Search of the Castaways & Jules Verne & Adventure fiction & English & Yes& 4 & 86\\
The Redeemer & Jo Nesbo & Crime fiction& English & No& 3 & 82\\
Drawing Sword & Du Liang & Historical fiction & Chinese & No& 3 & {70}\\
Dwelling Narrowness & Liu Liu & Social critique & Chinese & No& 3 & {92}\\
Distant Saviour & Dou Dou & Romance & Chinese & No& 4 & {48}\\
\midrule
Total & -- & -- & -- & -- & 40 & 795 \\
\quad English & -- & -- & English & -- & 13 & 319  \\
\quad Chinese & -- & -- & Chinese & -- & 25 & 476 \\
\quad Public & -- & -- & -- & Yes & 12 & 262 \\
\bottomrule
\end{tabular}
}
\vspace{-0.1in}
\caption{\small{Statistics of \datasetnamens.}}
\vspace{-0.1in}
\label{tab:statistics}
\end{table*}

\subsection{Dataset Construction}
\label{ssec:dataset_construction}

\paragraph{Label Definitions}
To facilitate human annotation, we categorize the consistent and contradictory cases into fine-grained types. Definitions and representative examples are provided in Table~\ref{tab:definitions}.

\paragraph{Guidelines for Human Annotation}
Annotators are instructed to follow the definitions and examples provided in Table~\ref{tab:definitions}. They are guided by a flowchart that first identify \emph{Contradict - Local}, \emph{Contradict - Global I}, and \emph{Contradict - Global II} respectively. If no contradiction is identified, they then determine whether the consistent prequel provides key missing information. During annotation, annotators are required to carefully consult the original book to identify any contradictions.

Annotation following this flowchart is generally sufficient for people familiar with the book. However, during trial annotation, we identified three issues hence introduced the following rules:

\begin{itemize}[noitemsep,nolistsep,leftmargin=*]
    \item First, judgments must be based solely on the content of the original novel. Adaptations, derivative works, or historical inspirations behind the characters should not be considered. Otherwise, annotators might incorrectly flag a contradiction on a case consistent with the novel based on, for example, a historical figure’s biography.
    \item Second, annotators should assume that the prequel is followed immediately by the original story, with no additional text or events in between. This rule addresses a tendency among human annotators to over-interpret. For instance, when facing a clear contradiction, they might imagine that some unseen event occurred between the prequel and the original story to make it eventually consistent. This rule reduces such subjectivity.
    \item Finally, unless a character's statements are later explicitly revealed to be deceptive, they should be treated as factual, akin to narrative exposition. Otherwise, one could dismiss any contradiction with the canonical text as intentional deceit.
\end{itemize}

\paragraph{Candidate Prequel Generation}
We prompt DeepSeek-R1 and GPT-4o to generate prequel samples, using the prompt provided in Appendix~\ref{app:prequel_generation}. In the prompt, we explicitly instruct the LLMs to generate a prequel for each character in Markdown list format.
Each bullet point is treated as an individual prequel example for annotators to label according to the types defined in Table~\ref{tab:definitions}.

\paragraph{Annotation Details}
Four annotators labeled the prequels for 40 characters across 13 books (see Appendix~\ref{app:dataset_char}), as shown in Table~\ref{tab:statistics}. These books were selected to represent diversity in genre, popularity, and original language.
Two annotators are graduate students majoring in literature or related fields, while the other two major in computer science.
Each annotator worked on books they were familiar with to ensure high-quality annotations.\footnote{\scriptsize By “familiar”, we require that the annotator has read the book multiple times and can recall the overall plot in reasonable detail.}
The label distribution in Table~\ref{tab:statistics2} shows that humans identified contradictions in nearly half of the generated prequels.

The annotation process resulted in a total of 795 instances, with each case taking approximately 10 minutes to complete.
After training, the annotators reached substantial agreement, with a Kappa score of 0.7828, though some subjectivity in interpretation remained. Most unresolved cases were due to differing interpretations of characters, ambiguities left by the original authors, or inherently fuzzy logic. Representative examples are shown in Appendix~\ref{app:qualitative}, \emph{Examples \ref{ex:fuzzy_logic}} and \emph{\ref{ex:subjective_interpretation}}.

\begin{table}
\small
\centering
\begin{tabular}{l|c||l|c}
\toprule
\bf Label  & \bf Count & \bf Label  & \bf Count \\
\midrule
Consistent & 434 & Contradict  & 361\\
\quad Core & 270 & \quad Local  & 94\\
\quad Irrelevant & 164 & \quad Global  & 267\\
\bottomrule
\end{tabular}
\vspace{-0.1in}
\caption{\small{Statistics of annotated labels.}}
\vspace{-0.1in}
\label{tab:statistics2}
\end{table}

\begin{table*}
\small
\centering
\resizebox{\columnwidth}{!}{
\begin{tabular}{@{}l||c|cc||cc|cc|c}
\toprule
& \multicolumn{3}{c||}{\bf Overall F1-Scores} & \multicolumn{5}{c}{\bf F1-Scores on Subsets} \\
\bf Model &\bf Macro-Avg &\bf Consistent &\bf Contradict & \bf Chinese-Set & \bf English-Set  & \bf R1-Set & \bf GPT-Set & \bf Public-Set\\
\midrule
Qwen2.5-72B & 55.7 & 66.4 & 45.0 & 54.8 & 57.1 & 54.7 & 56.8 & 59.7\\
\quad + RAG top-40 & 56.4 & 57.9 & 54.9 & 56.9 & 55.5 & 52.6 & 60.8 & 56.0\\
Qwen3-32B & 53.5 & 69.7 & 37.4 & 55.3 & 50.7 & 50.7 & 58.3 & 55.9\\ 
\quad + RAG top-40 & 61.3 & 67.3 & 55.4 & 61.7 & 60.7 & 59.2 & 64.8 & 64.1 \\
Qwen3-235B-A22B & 57.3 & \bf 70.1 & 44.5 & 54.9 & 60.9 & 53.9 & 62.9 & 59.1\\
\quad + RAG top-40 & 59.7 & 59.7 & 59.7 & 58.8 & 60.8 & 59.2 & 60.5 & 60.7 \\
DeepSeek-R1 & 61.3 & 69.2 & 53.4 & 60.3 & 62.7 & 63.9 & 56.2 & \bf 66.2\\
\quad + RAG top-40 &59.1 & 51.3 &  66.9 & 63.1 & 53.0 & 57.3 & 61.2 & 61.8\\
\midrule
GPT4o & 57.8 & 69.9 & 45.8 & 57.2 & 58.9 & 58.8 & 55.1 & 62.7\\
\quad + RAG top-40 & 62.9 & 61.2 & 64.6 & 62.3 & \bf 63.7 & 61.3 & 64.5 & 63.2 \\
o3-mini & 53.5 & 68.8 & 38.2 & 52.5 & 54.8 & 53.3 & 52.2 & 56.4 \\
\quad + RAG top-40 & 60.0 & 67.0 & 53.0 & 58.5 & 62.3 & 57.2 & 63.7 & 64.0 \\
Gemini-2.5-Flash & 61.8 & 62.4 & 61.1 & 66.4 & 54.9 & 59.6 & 64.6 & 60.5  \\
\quad + RAG top-40 & 57.8 & 48.9 & 66.8 & 62.8 & 50.2 & 57.1 & 58.3 & 52.0 \\
Gemini-2.5-Pro &\bf 65.1 & 61.4 & \bf 68.9 & \bf 67.1 & 62.2 & \bf 64.1 & \bf 66.4 & 62.0 \\
\quad + RAG top-40 & 60.7 & 53.7 & 67.8 & 61.0 & 60.2 & 58.4 & 63.6 & 59.9\\
\midrule
Humans & 81.7$^*$ & 79.5$^*$ & 83.9$^*$ &\multicolumn{5}{c}{--}\\
\bottomrule
\end{tabular}
}
\caption{\small{Comparison of different LLMs on the full set of \datasetnamens. *: Experiments conducted on a subset for reference.}}
\label{tab:overall_results}
\end{table*}

\section{Compared Methods}
We compare multiple state-of-the-art LLMs under the following settings. The implementation details can be found in Appendix~\ref{app:implementation_details}.
\paragraph{LLMs with Vanilla Few-Shot ICL}
This is the vanilla prompting approach that first presents the LLMs with the task instruction and then $k$ examples ($k=5$ in our case).
It does not provide book context in the input so the LLMs need to rely on their inherent parametric knowledge to solve the task.
We use the prompts shown in
Figure~\ref{fig:cn_rag_prompt_template} and \ref{fig:en_rag_prompt_template} from appendix, but with the field \emph{Original Excerpt} omitted from the input.

For open-source LLMs, we use the \emph{Instruct} versions of the Qwen2.5/3 models and DeepSeek-R1. We also compare against commercial LLMs, including GPT-4o, o3-mini, and the Gemini 2.5 family, accessed via API calls.

\paragraph{Retrieval-Augmented LLMs}
We enable the LLMs to access the canonical novels via retrieval-augmented generation (RAG)~\citep{rag,guu2020realm}. We experiment with various embedding models and hyperparameters, as detailed in Section~\ref{ssec:analysis}. Our final system uses the Qwen3-Embedding-8B model to retrieve the top 40 chunks, each with a length of 500 tokens.

\paragraph{In-Domain Post-Training}
Prior work has shown that fine-tuning with as few as 1K examples can elicit specific capabilities in LLMs, such as mathematical reasoning or general instruction following~\citep{zhou2023lima,zhao2024long,muennighoff2025s1}. These successes rely on the assumption that such capabilities are already present in the model acquired by pretraining thus can be activated with minimal supervision.

In contrast, if an LLM lacks the potential for a given capability, training with a small number of examples is unlikely to produce meaningful gains. Thus, the effectiveness of low-resource in-domain training can serve as a diagnostic tool to assess the intrinsic difficulty of a task, as demonstrated in~\citep{yu2025stochastic}.
Following this idea, we fine-tune on our labeled dataset (excluding the human study subset, $\sim$700 examples) and evaluate on the held-out human study subset.

\paragraph{Many-Shot ICL}
Similar to the in-domain training approach, many-shot ICL~\citep{agarwal2024manyshot,bertsch2025context} provides a large number of examples in the input context to elicit the latent capabilities of the LLM. We use the same data split as in the in-domain training experiment.

\paragraph{Commercial DeepResearch}

Commercial deep research services are offered by several companies, notably OpenAI DeepResearch\footnote{https://openai.com/index/introducing-deep-research/} and Google Gemini DeepResearch\footnote{https://gemini.google/overview/deep-research/}. These services showcase the ability to retrieve and synthesize information from multiple sources to generate reports using an agentic approach.
We use the web interface of OpenAI DeepResearch, which has demonstrated strong performance across a wide range of everyday tasks.

\section{Experimental Results}

\subsection{Human Performance}
\label{ssec:human_study}

We selected 100 examples to compute human performance. Three participants who had not involved in our task annotations and have  similar backgrounds to our annotators were asked to annotate examples from books they were familiar with.
The results show strong performance, with an F1 score of 81.7\% (an accuracy of 82\%), indicating that the task is largely solvable by humans.

Upon examining the disagreements, we found that most could be resolved, as they were often due to annotators overlooking information (either from fatigue after extended work or the the fallibility of human memory).
The remaining unresolved cases are of similar types to those analyzed in Section~\ref{ssec:dataset_construction}.

Our study further reveals that humans tend to adopt a DeepResearch-style approach, which involves iteratively generating hypotheses and resolving them by locating relevant supporting evidence.

\begin{table*}[t]
\small
\centering
\resizebox{1.\columnwidth}{!}{
\begin{tabular}{@{}l|l|c|c|cc|c|c}
\toprule
\multirow{2}{*}{\bf Methods} & \multirow{2}{*}{\bf Model} & \multirow{2}{*}{\bf Source} & \multicolumn{3}{c|}{\bf F1-Scores}&\multicolumn{2}{c}{\bf Accuracy} \\
&&&\bf Avg &\bf Consistent & \bf Contradict&\bf Answer& \bf Reason\\
\midrule
\multirow{7}{*}{No-Context} & Qwen3-32B & -- &53.8$_{\pm 1.6}$ & 63.2$_{\pm 1.7}$ & 44.4$_{\pm 1.5}$ & 55.7$_{\pm 1.7}$ & -- \\
& Qwen3-235B-A22B & -- & 57.4$_{\pm 3.5}$ & 62.1$_{\pm 3.3}$ & 52.8$_{\pm 3.9}$ & 57.7$_{\pm 3.1}$ & --\\
& DeepSeek-R1 & -- & \bf 65.0$_{\pm 0.7}$ & \bf 65.0$_{\pm 1.9}$ & 65.0$_{\pm 2.0}$ & 65.0$_{\pm 0.8}$& 47$^*$ \\
& GPT4o & -- & 58.8$_{\pm 1.1}$ & 64.7$_{\pm 1.5}$ & 52.9$_{\pm 1.4}$ & 59.7$_{\pm 1.2}$ &--\\
& o3-mini & -- & 50.0$_{\pm 2.0}$ &55.7$_{\pm 0.6}$ &44.3$_{\pm 3.7}$ &50.7$_{\pm 1.7}$ & --\\
& Gemini-2.5-Flash & -- & 63.4$_{\pm 3.7}$ &55.2$_{\pm 6.0}$ &71.6$_{\pm 1.6}$ &65.3$_{\pm 2.9}$ & -- \\
& Gemini-2.5-Pro & -- & 64.3$_{\pm 1.6}$ &52.0$_{\pm 2.8}$ &\bf 76.8$_{\pm 0.4}$ &\bf 68.7$_{\pm 0.9}$& 43$^*$\\

\midrule
\multirow{6}{*}{RAG} & Qwen3-32B & Book & 60.5$_{\pm 0.5}$ & \bf  60.0$_{\pm 2.1}$ & 61.0$_{\pm 3.2}$ & 60.7$_{\pm 0.5}$ & --\\
& Qwen3-235B-A22B & Book & \bf 63.1$_{\pm 2.5}$ &56.9$_{\pm 4.2}$ &69.3$_{\pm 2.3}$ &64.0$_{\pm 2.2}$ & 45$^*$ \\
\multirow{4}{*}{(top-40, total length: 20K)}& DeepSeek-R1 & Book & 59.1$_{\pm 4.1}$ & 42.4$_{\pm 6.7}$ & \bf 75.9$_{\pm 1.7}$ & 66.0$_{\pm 2.8}$ & 47$^*$\\
& GPT4o & Book &60.2$_{\pm 2.7}$ & 50.8$_{\pm 3.7}$ & 69.6$_{\pm 1.8}$ & 62.0$_{\pm 2.4}$ & -- \\
& o3-mini & Book & 55.7$_{\pm 1.9}$ &54.6$_{\pm 2.2}$ &56.7$_{\pm 1.7}$ &55.7$_{\pm 1.9}$ & -- \\
& Gemini-2.5-Flash & Book & 60.7$_{\pm 1.5}$ &45.8$_{\pm 3.3}$ &75.6$_{\pm 0.4}$ &\bf 66.3$_{\pm 0.5}$ & 47$^*$ \\
& Gemini-2.5-Pro & Book & 55.8$_{\pm 1.6}$ &37.5$_{\pm 2.5}$ &74.1$_{\pm 0.8}$ &63.3$_{\pm 1.2}$  & -- \\
\midrule
\multirow{2}{*}{In-Domain Training} & \multirow{2}{*}{Qwen3-32B} & -- & 52.4$_{\pm 1.5}$ & 58.7$_{\pm 3.0}$ & 46.1$_{\pm 1.5}$ & 53.3$_{\pm 1.9}$ & -- \\
&  & Book & 59.7$_{\pm 0.9}$ & 60.1$_{\pm 0.2}$ & 59.2$_{\pm 2.1}$ & 59.7$_{\pm 0.9}$ & --\\
\midrule
\multirow{2}{*}{Many-Shot ICL} & DeepSeek-R1 & -- &62.3$_{\pm 2.3}$ &62.0$_{\pm 1.9}$ &62.7$_{\pm 2.9}$ &62.3$_{\pm 2.4}$ & -- \\
& Gemini-2.5-Pro & -- & 59.5$_{\pm 1.4}$ &46.0$_{\pm 2.6}$ &73.0$_{\pm 0.3}$ &64.0$_{\pm 0.8}$ & -- \\
\midrule
OpenAI DeepResearch & o3 & Web& 62.5 & 58.4 & 66.7 & 63 & 51 \\
\midrule
Humans & -- & Book& 81.7$_{\pm 4.4}$ & 79.5$_{\pm 7.6}$ & 83.9$_{\pm 4.9}$ & 82$_{\pm 3.9}$ & 79\\
\bottomrule
\end{tabular}
}
\vspace{-0.1in}
\caption{\small{Comparing different systems on the human-study subset. $^*$: We conduct human verification on the most accurate results from the three runs.
For in-domain training experiments, we use the largest model (32B) supported on our infrastructure (8xA800). For many-shot ICL experiments, we compare with the two best-performed models in the \emph{No-Context} block.}}
\vspace{-0.1in}
\label{tab:human_study_set_results}
\end{table*}

\subsection{Results of LLMs}

\paragraph{Comparison across LLMs}
Table~\ref{tab:overall_results} compares state-of-the-art open-source and commercial LLMs on our task. The Gemini-2.5-Pro model shows a clear advantage over the others, yet still falls short of human performance by $>$15\%.

Another key finding is that, except for the Gemini-2.5 models, all other LLMs tend to overpredict the \emph{Consistent} label when not given access to the original books, resulting in unbalanced performance. This suggests that these models struggle to encode details of the novels within their parametric knowledge, underscoring the importance of incorporating canonical books in our task.

A per-book breakdown of the results from this study and the following RAG study is provided in Appendix~\ref{app:decomposition}.

\paragraph{Impact of RAG}
Table~\ref{tab:overall_results} also presents the performance of various LLMs under the RAG setting. For most models, RAG improves the F1 score on the \emph{Contradict} class. However, we also observe a tendency toward over-rejection, where LLMs predict \emph{Contradict} more frequently with hypercritical reasoning that focuses on minor or debatable inconsistencies.
One illustrative example can be found in Appendix~\ref{app:qualitative}, \emph{Example \ref{ex:rag_failure}}.

Notably, RAG results in worse performance for the Gemini-2.5-Pro model on both classes. This suggests that the retrieved contexts do not effectively contribute new or useful information for the strong Gemini-Pro model. It also highlights a broader limitation that despite recent advancements, long-context reasoning remains a persistent challenge for LLMs.

\paragraph{Results of In-Domain Training and Many-Shot ICL}
As shown in Table~\ref{tab:human_study_set_results}, on our held-out subset, neither in-domain fine-tuning nor many-shot ICL improves performance over the baseline usage of the same LLMs respectively. It indicates that current LLMs are still fundamentally limited in the type of reasoning required for our task.

\paragraph{Results of DeepResearch}
Finally, Table~\ref{tab:human_study_set_results} shows that DeepResearch performs worse than the best LLMs, both with and without RAG. It is also less effective at identifying contradictory evidence compared to most RAG-based systems. Since DeepResearch primarily relies on retrieving human-written analyses from the Internet, these results suggest that our task cannot be solved using existing external commentary or interpretations alone.

\subsection{Analysis}
\label{ssec:analysis}

\paragraph{Correct Answer with Incorrect Reasoning}
We manually verify the outputs of the LLMs and find that, although they correctly answer a large portion of the tasks, they often fail to arrive at the correct answers through valid reasoning.
Due to the labor-intensive nature of this evaluation, we select the best outputs from systems that achieve the highest performance on at least one metric in each block of Table~\ref{tab:human_study_set_results}. We also include DeepSeek-R1 with RAG, as its best run yields the highest answer accuracy.

\begin{figure}
\centering
\includegraphics[width=0.8\columnwidth]{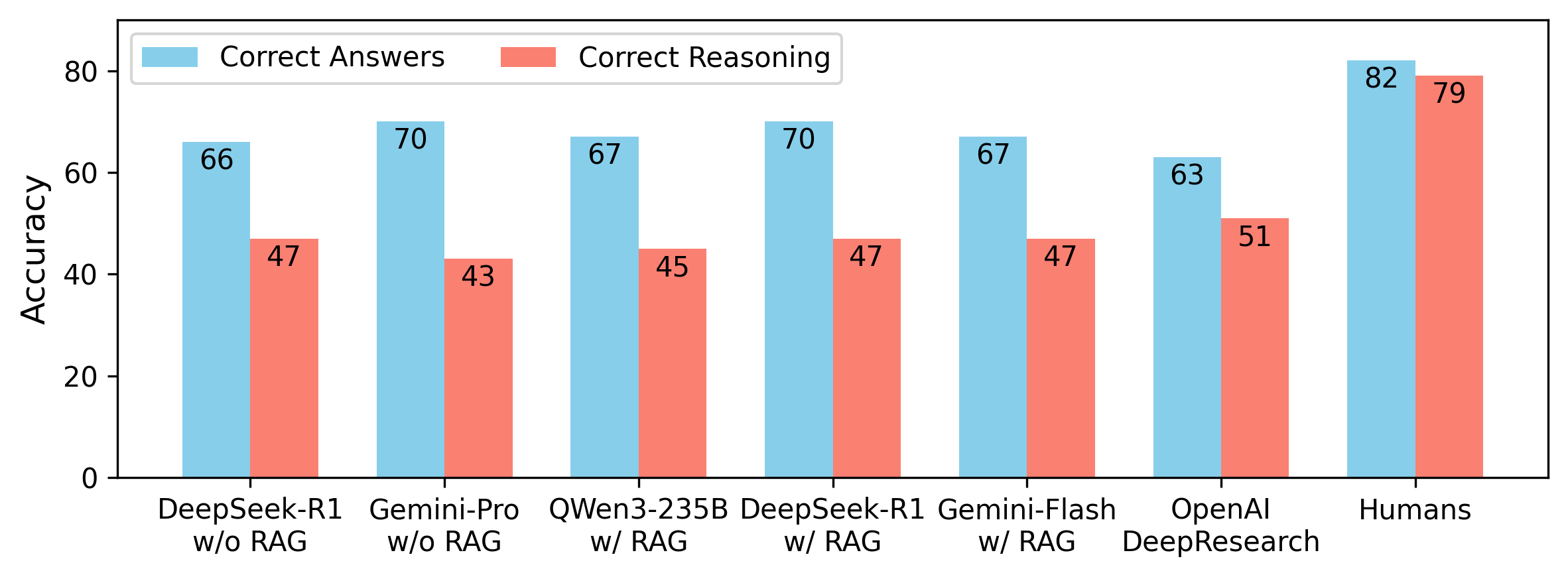}
\vspace{-0.2in}
  \caption{\small Accuracy the same methods when taking reasoning correctness into consideration.}
  \vspace{-0.1in}
  \label{fig:reasoning_accuracy}
\end{figure}

The final column of Table~\ref{tab:human_study_set_results} and Figure~\ref{fig:reasoning_accuracy} present the results, revealing a clear gap between answer accuracy and reasoning accuracy.
A such example can be found in Appendix~\ref{app:qualitative}, \emph{Example \ref{ex:incorrect_reason}}.
In contrast, human annotators generally agree on their reasoning, suggesting that current models still lack true comprehension in solving our task.

It is noteworthy that, despite its lower answer accuracy, DeepResearch exhibits the smallest performance drop when considering reasoning accuracy. This suggests combining a strong reasoning model with reflective mechanisms can lead to more reliable reasoning traces.

\begin{figure}[t] 
\centering
\includegraphics[width=0.4\columnwidth]{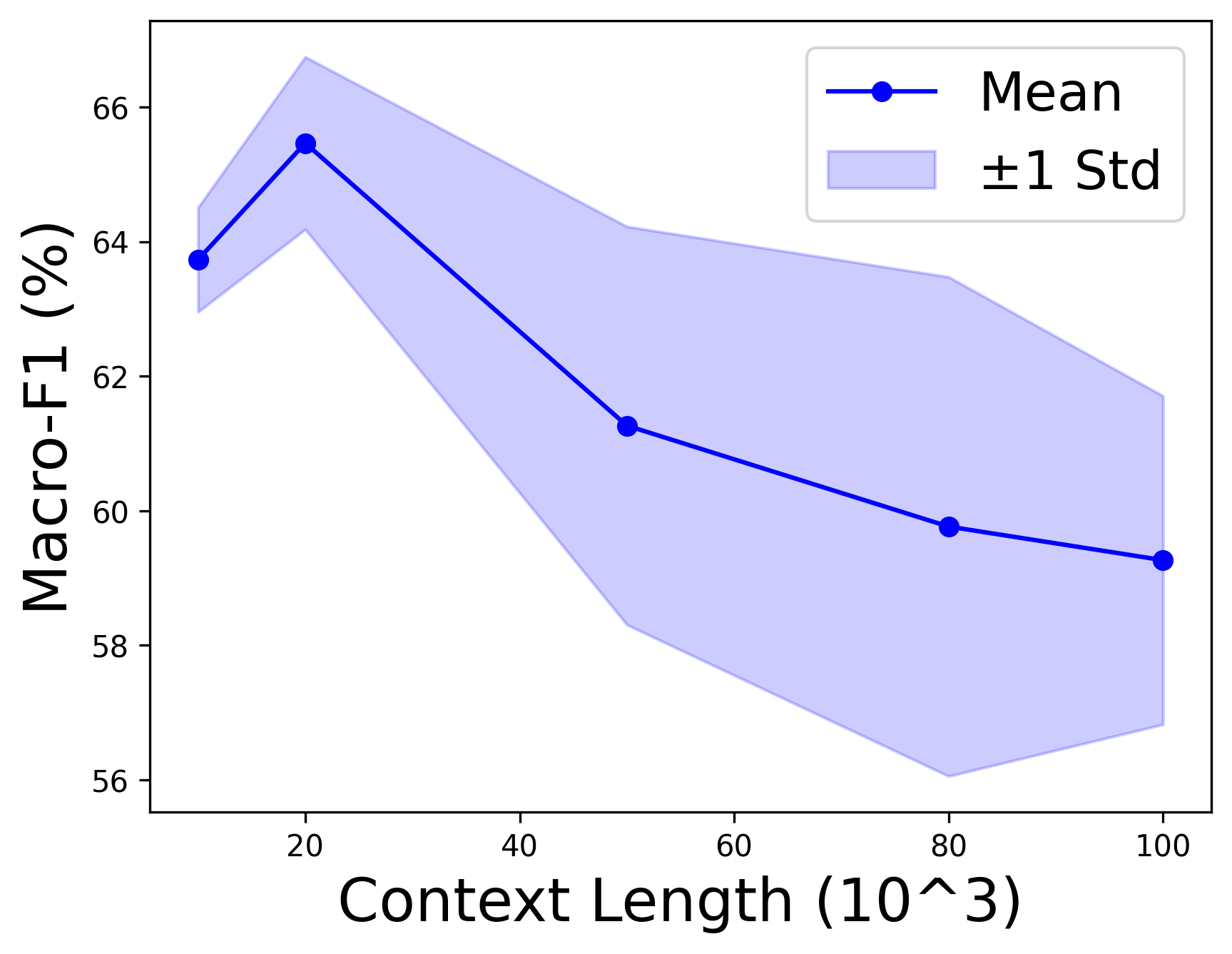}
\includegraphics[width=0.4\columnwidth]{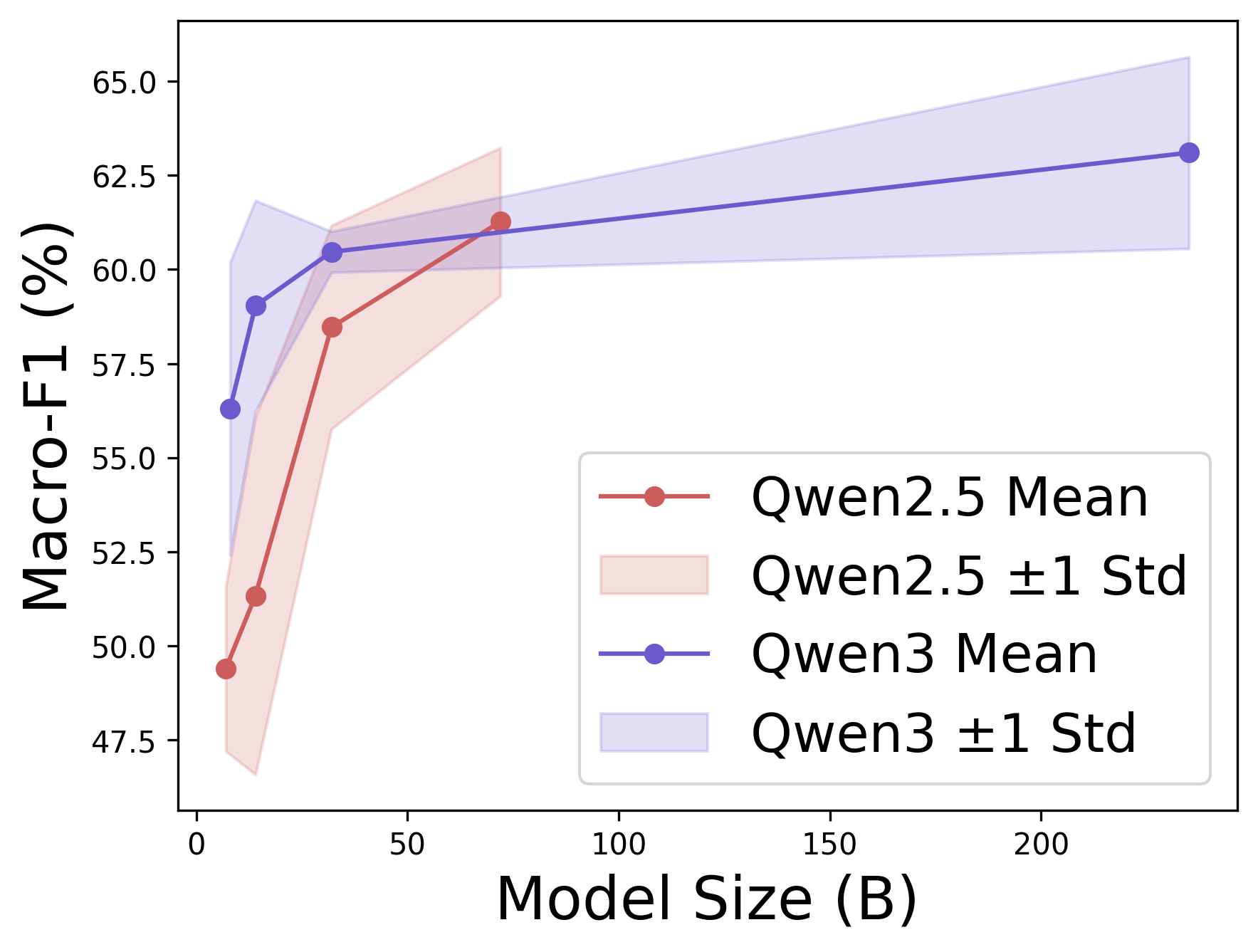}
\vspace{-0.1in}
  \caption{\small (a) Effect of retrieved context length on RAG performance. We use Qwen3-Embedding-8B and GPT4o to generate the results on the human study subset. (b) Study of the impact of model sizes under the best setting from (a).}
  \label{fig:effect_topk}
\end{figure}

\paragraph{Impact of Context Length in RAG}
Figure~\ref{fig:effect_topk}(a) investigates the impact of retrieved context lengths. The RAG system achieves its best performance when the input length is around 20k tokens.
With shorter contexts, the retriever often fails to include important evidence due to limited capacity. Conversely, with longer contexts, the excess information can overwhelm the LLM and hinder its ability to effectively utilize the extended context.

\paragraph{Different Retrieval Methods}
Table~\ref{tab:retrieval_comparison} provides ablation study on our retrieval method using the following variations:
\begin{itemize}[noitemsep,nolistsep,leftmargin=*]
\item \emph{Sorting the retrieved chunks according to their order in the books}: While this intuitively provides a more coherent context, it overlooks chunk relevance, resulting in decreased performance.
\item \emph{Doubling the chunk size while keeping the input length}: 
This causes a slight performance drop, likely due to the reduced effectiveness of embedding models on longer chunks~\citep{wu2025sitembv15improvedcontextawaredense}.
\item \emph{Replacing the embedding model with BGE-M3}: This slightly reduces performance. 
\item \emph{Replacing with BGE-M3 while doubling the chunk size}: This results in a further decrease in performance, likely due to BGE-M3's weaker handling of long inputs.
\end{itemize}

\begin{table}
\small
\centering
\begin{tabular}{@{}lcc||c|cc}
\toprule
& & & \multicolumn{3}{c}{\bf Overall F1-Scores} \\
\bf Retriever&\bf Top-k& \bf Len &\bf Macro-Avg &\bf Consis. &\bf Contra. \\
\midrule
No-RAG & -- & -- & 57.8 & \bf 69.9 & 45.8 \\
Qwen3-8B &40 & 500 & \bf 62.9 & 61.2 & \bf 64.6\\
\,\, + sorted  &40 & 500 & 60.1 & 58.8 & 61.5\\
Qwen3-8B & 20 &1k & 61.9 & 62.3 & 61.4\\
BGE-M3 &40 & 500 & 61.4 & 61.9 & 60.9 \\
BGE-M3 & 20 &1k  & 59.7 & 59.1 & 60.4\\
\bottomrule
\end{tabular}
\caption{\small{Comparison of different retrieval methods with the same context length. We use GPT4o to generate the results.}}
\label{tab:retrieval_comparison}
\end{table}

\paragraph{Effect of Model Scaling}
Figure~\ref{fig:effect_topk}(b) shows how model performance changes with increasing model size. We experiment with both the Qwen2.5 and Qwen3 series.
As shown in Table~\ref{tab:overall_results}, Qwen models rely on RAG to incorporate additional knowledge and improve performance. Therefore, all experiments in this section are conducted under the RAG setting, with each model evaluated over three runs.

The results indicate that for both Qwen series, performance consistently improves as model size increases. However, this improvement begins to plateau beyond the 32B model. Notably, the 235B model even outperforms the larger 671B R1 model on certain metrics, suggesting that simply scaling up model size is not efficient to our task.

\section{Discussions}

\paragraph{Limited Long Context Reasoning Capability in Recent LLMs}
Our results in Table~\ref{tab:overall_results} show that while some LLMs benefit from retrieved contexts, advanced reasoning models, such as DeepSeek-R1 and Gemini-2.5, exhibit a notable performance drop when context is provided. This gap becomes more pronounced as the base models grow stronger.

These findings suggest that recent improvements in LLMs' general reasoning capabilities do not necessarily translate to better long-context reasoning. One possible explanation is that as models become more powerful, their internal (parametric) knowledge is more efficient to solve tasks, making them prone to ignore external inputs during post-training. This highlights the need for improved training data and strategies specifically designed to encourage long-context reasoning.

\paragraph{Model Bias in Our Dataset Construction Method}
If our dataset construction method introduced bias, we would expect a model to perform worse on examples it generated itself, assuming that LLMs inherently trust their own outputs. However, the results in Table~\ref{tab:overall_results} show that this is not the case, indicating that our construction process does not introduce significant bias toward any particular model family.
This also suggests that state-of-the-art LLMs do not inherently trust their own generations and remain susceptible to hallucination, even when evaluating content they previously produced.

\paragraph{Measuring Fluid Intelligence in Natural Language Space}
Combined with the observation that DeepResearch performs poorly on our task, it becomes evident that the task cannot be solved simply by retrieving existing information from the web.
Instead, it requires generating new knowledge through reasoning based on learned rules, aligning with the notion of fluid intelligence tests \citep{chollet2019measure,chollet2025arc,yu2025stochastic}.
Unlike prior work, our task represents the first fluid intelligence assessment conducted entirely in the natural language space.

\section{Conclusion}
We introduce \datasetnamens, a new benchmark for evaluating long-context comprehension and reasoning in LLMs. Our task design addresses several key shortcuts present in prior long-context benchmarks.
Experiments show that state-of-the-art models still fall significantly short of human performance, particularly in generating valid reasoning. \datasetname calls for further research into robust long-context understanding and the development of models with stronger global reasoning capabilities.

\section*{Limitations}
Our task empirically mitigates the shortcuts observed in prior work (as shown in Figure~\ref{fig:intro}). However, due to the inherent complexity of long-context reasoning and the subjective nature of interpreting literary narratives, human performance on our task is also non-perfect. 
In future work, we aim to improve the annotation framework to further enhance inter-annotator consistency and reduce the annotation and human study workload of the task.

\bibliography{iclr2025_conference}

\begin{thebibliography}{66}
\providecommand{\natexlab}[1]{#1}
\providecommand{\url}[1]{\texttt{#1}}
\expandafter\ifx\csname urlstyle\endcsname\relax
  \providecommand{\doi}[1]{doi: #1}\else
  \providecommand{\doi}{doi: \begingroup \urlstyle{rm}\Url}\fi

\bibitem[Agarwal et~al.(2024)Agarwal, Singh, Zhang, Bohnet, Rosias, Chan, Zhang, Anand, Abbas, Nova, Co-Reyes, Chu, Behbahani, Faust, and Larochelle]{agarwal2024manyshot}
Rishabh Agarwal, Avi Singh, Lei~M Zhang, Bernd Bohnet, Luis Rosias, Stephanie~C.Y. Chan, Biao Zhang, Ankesh Anand, Zaheer Abbas, Azade Nova, John~D Co-Reyes, Eric Chu, Feryal Behbahani, Aleksandra Faust, and Hugo Larochelle.
\newblock Many-shot in-context learning.
\newblock In \emph{The Thirty-eighth Annual Conference on Neural Information Processing Systems}, 2024.
\newblock URL \url{https://openreview.net/forum?id=AB6XpMzvqH}.

\bibitem[Asai et~al.(2023)Asai, Wu, Wang, Sil, and Hajishirzi]{selfrag}
Akari Asai, Zeqiu Wu, Yizhong Wang, Avirup Sil, and Hannaneh Hajishirzi.
\newblock Self-rag: Learning to retrieve, generate, and critique through self-reflection.
\newblock In \emph{The Twelfth International Conference on Learning Representations}, 2023.

\bibitem[Bai et~al.(2024{\natexlab{a}})Bai, Lv, Zhang, Lyu, Tang, Huang, Du, Liu, Zeng, Hou, Dong, Tang, and Li]{DBLP:conf/acl/BaiLZL0HDLZHDTL24}
Yushi Bai, Xin Lv, Jiajie Zhang, Hongchang Lyu, Jiankai Tang, Zhidian Huang, Zhengxiao Du, Xiao Liu, Aohan Zeng, Lei Hou, Yuxiao Dong, Jie Tang, and Juanzi Li.
\newblock Longbench: {A} bilingual, multitask benchmark for long context understanding.
\newblock In \emph{Proceedings of the 62nd Annual Meeting of the Association for Computational Linguistics (Volume 1: Long Papers), {ACL} 2024, Bangkok, Thailand, August 11-16, 2024}, 2024{\natexlab{a}}.
\newblock URL \url{https://doi.org/10.18653/v1/2024.acl-long.172}.

\bibitem[Bai et~al.(2024{\natexlab{b}})Bai, Tu, Zhang, Peng, Wang, Lv, Cao, Xu, Hou, Dong, et~al.]{bai2024longbench}
Yushi Bai, Shangqing Tu, Jiajie Zhang, Hao Peng, Xiaozhi Wang, Xin Lv, Shulin Cao, Jiazheng Xu, Lei Hou, Yuxiao Dong, et~al.
\newblock Longbench v2: Towards deeper understanding and reasoning on realistic long-context multitasks.
\newblock \emph{arXiv preprint arXiv:2412.15204}, 2024{\natexlab{b}}.

\bibitem[Bertsch et~al.(2025)Bertsch, Ivgi, Xiao, Alon, Berant, Gormley, and Neubig]{bertsch2025context}
Amanda Bertsch, Maor Ivgi, Emily Xiao, Uri Alon, Jonathan Berant, Matthew~R Gormley, and Graham Neubig.
\newblock In-context learning with long-context models: An in-depth exploration.
\newblock In \emph{Proceedings of the 2025 Conference of the Nations of the Americas Chapter of the Association for Computational Linguistics: Human Language Technologies (Volume 1: Long Papers)}, pp.\  12119--12149, 2025.

\bibitem[Chang et~al.(2023)Chang, Lo, Goyal, and Iyyer]{chang2023booookscore}
Yapei Chang, Kyle Lo, Tanya Goyal, and Mohit Iyyer.
\newblock Booookscore: A systematic exploration of book-length summarization in the era of llms.
\newblock \emph{arXiv preprint arXiv:2310.00785}, 2023.

\bibitem[Chollet(2019)]{chollet2019measure}
Fran{\c{c}}ois Chollet.
\newblock On the measure of intelligence.
\newblock \emph{arXiv preprint arXiv:1911.01547}, 2019.

\bibitem[Chollet et~al.(2025)Chollet, Knoop, Kamradt, Landers, and Pinkard]{chollet2025arc}
Francois Chollet, Mike Knoop, Gregory Kamradt, Bryan Landers, and Henry Pinkard.
\newblock Arc-agi-2: A new challenge for frontier ai reasoning systems.
\newblock \emph{arXiv preprint arXiv:2505.11831}, 2025.

\bibitem[Del{\'e}tang et~al.(2023)Del{\'e}tang, Ruoss, Duquenne, Catt, Genewein, Mattern, Grau-Moya, Wenliang, Aitchison, Orseau, et~al.]{deletang2023language}
Gr{\'e}goire Del{\'e}tang, Anian Ruoss, Paul-Ambroise Duquenne, Elliot Catt, Tim Genewein, Christopher Mattern, Jordi Grau-Moya, Li~Kevin Wenliang, Matthew Aitchison, Laurent Orseau, et~al.
\newblock Language modeling is compression.
\newblock \emph{arXiv preprint arXiv:2309.10668}, 2023.

\bibitem[Dong et~al.(2024)Dong, Tang, Li, Zhao, and Wen]{dong-etal-2024-bamboo}
Zican Dong, Tianyi Tang, Junyi Li, Wayne~Xin Zhao, and Ji-Rong Wen.
\newblock {BAMBOO}: A comprehensive benchmark for evaluating long text modeling capacities of large language models.
\newblock In Nicoletta Calzolari, Min-Yen Kan, Veronique Hoste, Alessandro Lenci, Sakriani Sakti, and Nianwen Xue (eds.), \emph{Proceedings of the 2024 Joint International Conference on Computational Linguistics, Language Resources and Evaluation (LREC-COLING 2024)}, pp.\  2086--2099, Torino, Italia, May 2024. ELRA and ICCL.
\newblock URL \url{https://aclanthology.org/2024.lrec-main.188/}.

\bibitem[Edge et~al.(2024)Edge, Trinh, Cheng, Bradley, Chao, Mody, Truitt, Metropolitansky, Ness, and Larson]{graphrag}
Darren Edge, Ha~Trinh, Newman Cheng, Joshua Bradley, Alex Chao, Apurva Mody, Steven Truitt, Dasha Metropolitansky, Robert~Osazuwa Ness, and Jonathan Larson.
\newblock From local to global: A graph rag approach to query-focused summarization.
\newblock \emph{arXiv preprint arXiv:2404.16130}, 2024.

\bibitem[Fang et~al.(2024)Fang, Wang, Liu, Zhang, Jegelka, Gao, Ding, and Wang]{fang2024wrong}
Lizhe Fang, Yifei Wang, Zhaoyang Liu, Chenheng Zhang, Stefanie Jegelka, Jinyang Gao, Bolin Ding, and Yisen Wang.
\newblock What is wrong with perplexity for long-context language modeling?
\newblock \emph{arXiv preprint arXiv:2410.23771}, 2024.

\bibitem[{Google DeepMind}(2024)]{notebooklm}
{Google DeepMind}.
\newblock {NotebookLM: Your AI-Powered Research Assistant}.
\newblock \url{https://notebooklm.google/}, 2024.
\newblock Accessed: 2025-07-24.

\bibitem[Gutierrez et~al.(2024)Gutierrez, Shu, Gu, Yasunaga, and Su]{hipporag}
Bernal~Jimenez Gutierrez, Yiheng Shu, Yu~Gu, Michihiro Yasunaga, and Yu~Su.
\newblock Hipporag: Neurobiologically inspired long-term memory for large language models.
\newblock \emph{Advances in Neural Information Processing Systems}, 37:\penalty0 59532--59569, 2024.

\bibitem[Guu et~al.(2020)Guu, Lee, Tung, Pasupat, and Chang]{guu2020realm}
Kelvin Guu, Kenton Lee, Zora Tung, Panupong Pasupat, and Ming-Wei Chang.
\newblock Realm: Retrieval-augmented language model pre-training.
\newblock \emph{arXiv preprint arXiv:2002.08909}, 2020.

\bibitem[Haosheng~Zou \& Zhang(2024)Haosheng~Zou and Zhang]{360-llama-factory}
Shousheng~Jia Haosheng~Zou, Xiaowei~Lv and Xiangzheng Zhang.
\newblock 360-llama-factory, 2024.
\newblock URL \url{https://github.com/Qihoo360/360-LLaMA-Factory}.

\bibitem[Hsieh et~al.(2024)Hsieh, Sun, Kriman, Acharya, Rekesh, Jia, Zhang, and Ginsburg]{hsieh2024ruler}
Cheng-Ping Hsieh, Simeng Sun, Samuel Kriman, Shantanu Acharya, Dima Rekesh, Fei Jia, Yang Zhang, and Boris Ginsburg.
\newblock Ruler: What's the real context size of your long-context language models?
\newblock \emph{arXiv preprint arXiv:2404.06654}, 2024.

\bibitem[Hu et~al.(2021)Hu, Wallis, Allen-Zhu, Li, Wang, Wang, Chen, et~al.]{hu2021lora}
Edward~J Hu, Phillip Wallis, Zeyuan Allen-Zhu, Yuanzhi Li, Shean Wang, Lu~Wang, Weizhu Chen, et~al.
\newblock Lora: Low-rank adaptation of large language models.
\newblock In \emph{International Conference on Learning Representations}, 2021.

\bibitem[Hu et~al.(2025)Hu, Zhou, Fan, Nie, Xia, Sun, Ye, Jin, Li, Chen, Zhang, Wang, Ye, Ghanem, Luo, and Li]{hu2025owl}
Mengkang Hu, Yuhang Zhou, Wendong Fan, Yuzhou Nie, Bowei Xia, Tao Sun, Ziyu Ye, Zhaoxuan Jin, Yingru Li, Qiguang Chen, Zeyu Zhang, Yifeng Wang, Qianshuo Ye, Bernard Ghanem, Ping Luo, and Guohao Li.
\newblock Owl: Optimized workforce learning for general multi-agent assistance in real-world task automation, 2025.
\newblock URL \url{https://arxiv.org/abs/2505.23885}.

\bibitem[Jimenez et~al.(2024)Jimenez, Yang, Wettig, Yao, Pei, Press, and Narasimhan]{jimenezswe24}
Carlos~E Jimenez, John Yang, Alexander Wettig, Shunyu Yao, Kexin Pei, Ofir Press, and Karthik~R Narasimhan.
\newblock Swe-bench: Can language models resolve real-world github issues?
\newblock In \emph{The Twelfth International Conference on Learning Representations, ICLR 2024}, 2024.

\bibitem[Kamradt(2023)]{kamradt2023needle}
Gregory Kamradt.
\newblock Needle in a haystack - pressure testing llms.
\newblock \url{https://github.com/gkamradt/LLMTest NeedleInAHaystack/tree/main}, 2023.
\newblock Accessed: 2025-07-23.

\bibitem[Karpinska et~al.(2024)Karpinska, Thai, Lo, Goyal, and Iyyer]{DBLP:conf/emnlp/KarpinskaTLGI24}
Marzena Karpinska, Katherine Thai, Kyle Lo, Tanya Goyal, and Mohit Iyyer.
\newblock One thousand and one pairs: {A} "novel" challenge for long-context language models.
\newblock In \emph{Proceedings of the 2024 Conference on Empirical Methods in Natural Language Processing, {EMNLP} 2024, Miami, FL, USA, November 12-16, 2024}, 2024.
\newblock URL \url{https://doi.org/10.18653/v1/2024.emnlp-main.948}.

\bibitem[Ko{\v{c}}isk{\`y} et~al.(2018)Ko{\v{c}}isk{\`y}, Schwarz, Blunsom, Dyer, Hermann, Melis, and Grefenstette]{kovcisky2018narrativeqa}
Tom{\'a}{\v{s}} Ko{\v{c}}isk{\`y}, Jonathan Schwarz, Phil Blunsom, Chris Dyer, Karl~Moritz Hermann, G{\'a}bor Melis, and Edward Grefenstette.
\newblock The narrativeqa reading comprehension challenge.
\newblock \emph{Transactions of the Association for Computational Linguistics}, 6:\penalty0 317--328, 2018.
\newblock URL \url{https://aclanthology.org/Q18-1023.pdf}.

\bibitem[Kuratov et~al.(2024)Kuratov, Bulatov, Anokhin, Rodkin, Sorokin, Sorokin, and Burtsev]{kuratov2024babilong}
Yuri Kuratov, Aydar Bulatov, Petr Anokhin, Ivan Rodkin, Dmitry~Igorevich Sorokin, Artyom Sorokin, and Mikhail Burtsev.
\newblock {BABIL}ong: Testing the limits of {LLM}s with long context reasoning-in-a-haystack.
\newblock In \emph{The Thirty-eight Conference on Neural Information Processing Systems Datasets and Benchmarks Track}, 2024.
\newblock URL \url{https://openreview.net/forum?id=u7m2CG84BQ}.

\bibitem[Lee et~al.(2025)Lee, Yoon, Jang, Lee, Song, Kim, and Kang]{lee-etal-2025-ethic}
Taewhoo Lee, Chanwoong Yoon, Kyochul Jang, Donghyeon Lee, Minju Song, Hyunjae Kim, and Jaewoo Kang.
\newblock {ETHIC}: Evaluating large language models on long-context tasks with high information coverage.
\newblock In Luis Chiruzzo, Alan Ritter, and Lu~Wang (eds.), \emph{Proceedings of the 2025 Conference of the Nations of the Americas Chapter of the Association for Computational Linguistics: Human Language Technologies (Volume 1: Long Papers)}, pp.\  5497--5512, Albuquerque, New Mexico, April 2025. Association for Computational Linguistics.
\newblock ISBN 979-8-89176-189-6.
\newblock \doi{10.18653/v1/2025.naacl-long.283}.
\newblock URL \url{https://aclanthology.org/2025.naacl-long.283/}.

\bibitem[Lewis et~al.(2020)Lewis, Perez, Piktus, et~al.]{rag}
Patrick Lewis, Ethan Perez, Aleksandra Piktus, et~al.
\newblock Retrieval-augmented generation for knowledge-intensive nlp tasks.
\newblock \emph{Advances in Neural Information Processing Systems}, 33:\penalty0 9459--9474, 2020.

\bibitem[Li et~al.(2024)Li, Wang, Zheng, and Zhang]{DBLP:conf/acl/LiWZZ24}
Jiaqi Li, Mengmeng Wang, Zilong Zheng, and Muhan Zhang.
\newblock Loogle: Can long-context language models understand long contexts?
\newblock In \emph{Proceedings of the 62nd Annual Meeting of the Association for Computational Linguistics (Volume 1: Long Papers), {ACL} 2024, Bangkok, Thailand, August 11-16, 2024}, 2024.
\newblock URL \url{https://doi.org/10.18653/v1/2024.acl-long.859}.

\bibitem[Li et~al.(2025{\natexlab{a}})Li, Zhang, Jiang, Xie, Huang, Wang, and Cheng]{li2025lara}
Kuan Li, Liwen Zhang, Yong Jiang, Pengjun Xie, Fei Huang, Shuai Wang, and Minhao Cheng.
\newblock Lara: Benchmarking retrieval-augmented generation and long-context llms--no silver bullet for lc or rag routing.
\newblock \emph{arXiv preprint arXiv:2502.09977}, 2025{\natexlab{a}}.

\bibitem[Li et~al.(2025{\natexlab{b}})Li, Zhang, Zhang, Duan, Liu, and Chen]{li2025needlebenchllmsretrievalreasoning}
Mo~Li, Songyang Zhang, Taolin Zhang, Haodong Duan, Yunxin Liu, and Kai Chen.
\newblock Needlebench: Can llms do retrieval and reasoning in information-dense context?, 2025{\natexlab{b}}.
\newblock URL \url{https://arxiv.org/abs/2407.11963}.

\bibitem[Li et~al.(2025{\natexlab{c}})Li, Zhang, Do, Yue, and Chen]{li2025longcontext}
Tianle Li, Ge~Zhang, Quy~Duc Do, Xiang Yue, and Wenhu Chen.
\newblock Long-context {LLM}s struggle with long in-context learning.
\newblock \emph{Transactions on Machine Learning Research}, 2025{\natexlab{c}}.
\newblock ISSN 2835-8856.
\newblock URL \url{https://openreview.net/forum?id=Cw2xlg0e46}.

\bibitem[Liu et~al.(2024)Liu, Lin, Hewitt, Paranjape, Bevilacqua, Petroni, and Liang]{liu2024lost}
Nelson~F Liu, Kevin Lin, John Hewitt, Ashwin Paranjape, Michele Bevilacqua, Fabio Petroni, and Percy Liang.
\newblock Lost in the middle: How language models use long contexts.
\newblock \emph{Transactions of the Association for Computational Linguistics}, 12, 2024.

\bibitem[Muennighoff et~al.(2025)Muennighoff, Yang, Shi, Li, Fei-Fei, Hajishirzi, Zettlemoyer, Liang, Cand{\`e}s, and Hashimoto]{muennighoff2025s1}
Niklas Muennighoff, Zitong Yang, Weijia Shi, Xiang~Lisa Li, Li~Fei-Fei, Hannaneh Hajishirzi, Luke Zettlemoyer, Percy Liang, Emmanuel Cand{\`e}s, and Tatsunori Hashimoto.
\newblock s1: Simple test-time scaling.
\newblock \emph{arXiv preprint arXiv:2501.19393}, 2025.

\bibitem[Ni et~al.(2024)Ni, Cai, Wei, Wang, Yin, and Li]{ni2024xl}
Xuanfan Ni, Hengyi Cai, Xiaochi Wei, Shuaiqiang Wang, Dawei Yin, and Piji Li.
\newblock Xl$^{2}$ bench: A benchmark for extremely long context understanding with long-range dependencies.
\newblock \emph{arXiv preprint arXiv:2404.05446}, 2024.

\bibitem[OpenAI(2025)]{openai2025deepresearch}
OpenAI.
\newblock Deep research system card.
\newblock Technical report, OpenAI, 2025.

\bibitem[Press et~al.(2023)Press, Zhang, Min, Schmidt, Smith, and Lewis]{press2023measuring}
Ofir Press, Muru Zhang, Sewon Min, Ludwig Schmidt, Noah~A Smith, and Mike Lewis.
\newblock Measuring and narrowing the compositionality gap in language models.
\newblock In \emph{Findings of the Association for Computational Linguistics: EMNLP 2023}, pp.\  5687--5711, 2023.

\bibitem[Qiu et~al.(2024)Qiu, Li, Huang, Jiao, Zhong, and King]{DBLP:conf/emnlp/QiuLHJZK24}
Zexuan Qiu, Jingjing Li, Shijue Huang, Xiaoqi Jiao, Wanjun Zhong, and Irwin King.
\newblock Clongeval: {A} chinese benchmark for evaluating long-context large language models.
\newblock In \emph{Findings of the Association for Computational Linguistics: {EMNLP} 2024, Miami, Florida, USA, November 12-16, 2024}, 2024.
\newblock URL \url{https://doi.org/10.18653/v1/2024.findings-emnlp.230}.

\bibitem[Schlichtkrull et~al.(2023)Schlichtkrull, Guo, and Vlachos]{schlichtkrull2023averitec}
Michael Schlichtkrull, Zhijiang Guo, and Andreas Vlachos.
\newblock Averitec: A dataset for real-world claim verification with evidence from the web.
\newblock \emph{Advances in Neural Information Processing Systems}, 36:\penalty0 65128--65167, 2023.

\bibitem[Sutskever(2023)]{ilya2023}
Ilya Sutskever.
\newblock A theory of unsupervised learning, 2023.
\newblock URL \url{https://www.youtube.com/watch?v=AKMuA_TVz3A}.

\bibitem[Thorne et~al.(2018)Thorne, Vlachos, Christodoulopoulos, and Mittal]{thorne2018fever}
James Thorne, Andreas Vlachos, Christos Christodoulopoulos, and Arpit Mittal.
\newblock Fever: a large-scale dataset for fact extraction and verification.
\newblock \emph{arXiv preprint arXiv:1803.05355}, 2018.

\bibitem[Tirumala et~al.(2022)Tirumala, Markosyan, Zettlemoyer, and Aghajanyan]{tirumala2022memorization}
Kushal Tirumala, Aram Markosyan, Luke Zettlemoyer, and Armen Aghajanyan.
\newblock Memorization without overfitting: Analyzing the training dynamics of large language models.
\newblock \emph{Advances in Neural Information Processing Systems}, 35:\penalty0 38274--38290, 2022.

\bibitem[Wadden et~al.(2020)Wadden, Lin, Lo, Wang, van Zuylen, Cohan, and Hajishirzi]{wadden2020fact}
David Wadden, Shanchuan Lin, Kyle Lo, Lucy~Lu Wang, Madeleine van Zuylen, Arman Cohan, and Hannaneh Hajishirzi.
\newblock Fact or fiction: Verifying scientific claims.
\newblock \emph{arXiv preprint arXiv:2004.14974}, 2020.

\bibitem[Wang et~al.(2024{\natexlab{a}})Wang, Duan, Zhang, Lin, and Chen]{wang-etal-2024-ada}
Chonghua Wang, Haodong Duan, Songyang Zhang, Dahua Lin, and Kai Chen.
\newblock {A}da-{LE}val: Evaluating long-context {LLM}s with length-adaptable benchmarks.
\newblock In Kevin Duh, Helena Gomez, and Steven Bethard (eds.), \emph{Proceedings of the 2024 Conference of the North American Chapter of the Association for Computational Linguistics: Human Language Technologies (Volume 1: Long Papers)}, pp.\  3712--3724, Mexico City, Mexico, June 2024{\natexlab{a}}. Association for Computational Linguistics.
\newblock \doi{10.18653/v1/2024.naacl-long.205}.
\newblock URL \url{https://aclanthology.org/2024.naacl-long.205/}.

\bibitem[Wang et~al.(2024{\natexlab{b}})Wang, Ning, Pan, Wu, Guo, Deng, Bao, Wang, and Zhang]{DBLP:journals/corr/abs-2403-12766}
Cunxiang Wang, Ruoxi Ning, Boqi Pan, Tonghui Wu, Qipeng Guo, Cheng Deng, Guangsheng Bao, Qian Wang, and Yue Zhang.
\newblock Novelqa: A benchmark for long-range novel question answering.
\newblock \emph{CoRR}, abs/2403.12766, 2024{\natexlab{b}}.
\newblock URL \url{https://doi.org/10.48550/arXiv.2403.12766}.

\bibitem[Wang et~al.(2024{\natexlab{c}})Wang, Chen, Cheng, Liao, Zhang, Wu, Yu, Xu, Zhang, Luo, Li, Yang, Huang, and Li]{wang-etal-2024-leave}
Minzheng Wang, Longze Chen, Fu~Cheng, Shengyi Liao, Xinghua Zhang, Bingli Wu, Haiyang Yu, Nan Xu, Lei Zhang, Run Luo, Yunshui Li, Min Yang, Fei Huang, and Yongbin Li.
\newblock Leave no document behind: Benchmarking long-context {LLM}s with extended multi-doc {QA}.
\newblock In Yaser Al-Onaizan, Mohit Bansal, and Yun-Nung Chen (eds.), \emph{Proceedings of the 2024 Conference on Empirical Methods in Natural Language Processing}, pp.\  5627--5646, Miami, Florida, USA, November 2024{\natexlab{c}}. Association for Computational Linguistics.
\newblock \doi{10.18653/v1/2024.emnlp-main.322}.
\newblock URL \url{https://aclanthology.org/2024.emnlp-main.322/}.

\bibitem[Wang et~al.(2024{\natexlab{d}})Wang, Chen, Cheng, Liao, Zhang, Wu, Yu, Xu, Zhang, Luo, et~al.]{wang2024leave}
Minzheng Wang, Longze Chen, Fu~Cheng, Shengyi Liao, Xinghua Zhang, Bingli Wu, Haiyang Yu, Nan Xu, Lei Zhang, Run Luo, et~al.
\newblock Leave no document behind: Benchmarking long-context llms with extended multi-doc qa.
\newblock In \emph{Proceedings of the 2024 Conference on Empirical Methods in Natural Language Processing}, pp.\  5627--5646, 2024{\natexlab{d}}.

\bibitem[Wang et~al.(2025)Wang, Wang, Wang, Zhang, Li, Yang, Jin, Yu, Nguyen, Liu, et~al.]{wang2025ragen}
Zihan Wang, Kangrui Wang, Qineng Wang, Pingyue Zhang, Linjie Li, Zhengyuan Yang, Xing Jin, Kefan Yu, Minh~Nhat Nguyen, Licheng Liu, et~al.
\newblock Ragen: Understanding self-evolution in llm agents via multi-turn reinforcement learning.
\newblock \emph{arXiv preprint arXiv:2504.20073}, 2025.

\bibitem[Wu et~al.(2024)Wu, Wang, Yu, Zhang, Chang, and Yu]{wu2024longmemeval}
Di~Wu, Hongwei Wang, Wenhao Yu, Yuwei Zhang, Kai-Wei Chang, and Dong Yu.
\newblock Longmemeval: Benchmarking chat assistants on long-term interactive memory.
\newblock \emph{arXiv preprint arXiv:2410.10813}, 2024.

\bibitem[Wu et~al.(2025{\natexlab{a}})Wu, Li, Li, Liu, Xu, Li, Yeung, Zhou, and Yu]{wu2025sitembv15improvedcontextawaredense}
Junjie Wu, Jiangnan Li, Yuqing Li, Lemao Liu, Liyan Xu, Jiwei Li, Dit-Yan Yeung, Jie Zhou, and Mo~Yu.
\newblock Sitemb-v1.5: Improved context-aware dense retrieval for semantic association and long story comprehension, 2025{\natexlab{a}}.
\newblock URL \url{https://arxiv.org/abs/2508.01959}.

\bibitem[Wu et~al.(2025{\natexlab{b}})Wu, Bai, Hu, Tu, Hee, Li, and Lee]{wu2025shifting}
Yuhao Wu, Yushi Bai, Zhiqing Hu, Shangqing Tu, Ming~Shan Hee, Juanzi Li, and Roy Ka-Wei Lee.
\newblock Shifting long-context llms research from input to output.
\newblock \emph{arXiv preprint arXiv:2503.04723}, 2025{\natexlab{b}}.

\bibitem[Xiong et~al.(2023)Xiong, Liu, Molybog, Zhang, Bhargava, Hou, Martin, Rungta, Sankararaman, Oguz, Khabsa, Fang, Mehdad, Narang, Malik, Fan, Bhosale, Edunov, Lewis, Wang, and Ma]{long-llama}
Wenhan Xiong, Jingyu Liu, Igor Molybog, Hejia Zhang, Prajjwal Bhargava, Rui Hou, Louis Martin, Rashi Rungta, Karthik~Abinav Sankararaman, Barlas Oguz, Madian Khabsa, Han Fang, Yashar Mehdad, Sharan Narang, Kshitiz Malik, Angela Fan, Shruti Bhosale, Sergey Edunov, Mike Lewis, Sinong Wang, and Hao Ma.
\newblock Effective long-context scaling of foundation models, 2023.

\bibitem[Xiong et~al.(2021)Xiong, Zeng, Chakraborty, Tan, Fung, Li, and Singh]{xiong2021nystromformer}
Yunyang Xiong, Zhanpeng Zeng, Rudrasis Chakraborty, Mingxing Tan, Glenn Fung, Yin Li, and Vikas Singh.
\newblock Nystr{\"o}mformer: A nystr{\"o}m-based algorithm for approximating self-attention.
\newblock In \emph{Proceedings of the AAAI Conference on Artificial Intelligence}, volume~35, pp.\  14138--14148, 2021.
\newblock URL \url{https://ojs.aaai.org/index.php/AAAI/article/view/17664}.

\bibitem[Xu et~al.(2024{\natexlab{a}})Xu, Li, Yu, and Zhou]{xu2024narco}
Liyan Xu, Jiangnan Li, Mo~Yu, and Jie Zhou.
\newblock Fine-grained modeling of narrative context: A coherence perspective via retrospective questions.
\newblock In \emph{Proceedings of the 62nd Annual Meeting of the Association for Computational Linguistics (Volume 1: Long Papers)}, pp.\  5822--5838, 2024{\natexlab{a}}.

\bibitem[Xu \& Ma(2025)Xu and Ma]{xu2025llm}
Nan Xu and Xuezhe Ma.
\newblock Llm the genius paradox: A linguistic and math expert’s struggle with simple word-based counting problems.
\newblock In \emph{Proceedings of the 2025 Conference of the Nations of the Americas Chapter of the Association for Computational Linguistics: Human Language Technologies (Volume 1: Long Papers)}, pp.\  3344--3370, 2025.

\bibitem[Xu et~al.(2025{\natexlab{a}})Xu, Mei, Gao, Tan, Liang, and Zhang]{xu2025mem}
Wujiang Xu, Kai Mei, Hang Gao, Juntao Tan, Zujie Liang, and Yongfeng Zhang.
\newblock A-mem: Agentic memory for llm agents.
\newblock \emph{arXiv preprint arXiv:2502.12110}, 2025{\natexlab{a}}.

\bibitem[Xu et~al.(2024{\natexlab{b}})Xu, Ye, and Ren]{xu2024stresstesting}
Xiaoyue Xu, Qinyuan Ye, and Xiang Ren.
\newblock Stress-testing long-context language models with lifelong {ICL} and task haystack.
\newblock In \emph{The Thirty-eight Conference on Neural Information Processing Systems Datasets and Benchmarks Track}, 2024{\natexlab{b}}.
\newblock URL \url{https://openreview.net/forum?id=j6PTT6NB2O}.

\bibitem[Xu et~al.(2025{\natexlab{b}})Xu, Ye, Liu, Liu, Sun, Liu, Guo, Li, Liu, Huang, and Qiu]{xu2025detectiveqa}
Zhe Xu, Jiasheng Ye, Xiaoran Liu, Xiangyang Liu, Tianxiang Sun, Zhigeng Liu, Qipeng Guo, Linlin Li, Qun Liu, Xuanjing Huang, and Xipeng Qiu.
\newblock Detective{QA}: Evaluating long-context reasoning on detective novels.
\newblock In \emph{Workshop on Reasoning and Planning for Large Language Models}, 2025{\natexlab{b}}.
\newblock URL \url{https://openreview.net/forum?id=9ExIs5ELlk}.

\bibitem[Yang et~al.(2018)Yang, Qi, Zhang, Bengio, Cohen, Salakhutdinov, and Manning]{yang2018hotpotqa}
Zhilin Yang, Peng Qi, Saizheng Zhang, Yoshua Bengio, William Cohen, Ruslan Salakhutdinov, and Christopher~D Manning.
\newblock Hotpotqa: A dataset for diverse, explainable multi-hop question answering.
\newblock In \emph{Proceedings of EMNLP 2018}, pp.\  2369--2380, 2018.

\bibitem[Yen et~al.(2024)Yen, Gao, Hou, Ding, Fleischer, Izsak, Wasserblat, and Chen]{yen2024helmet}
Howard Yen, Tianyu Gao, Minmin Hou, Ke~Ding, Daniel Fleischer, Peter Izsak, Moshe Wasserblat, and Danqi Chen.
\newblock Helmet: How to evaluate long-context language models effectively and thoroughly.
\newblock \emph{arXiv preprint arXiv:2410.02694}, 2024.

\bibitem[Yin et~al.(2021)Yin, Radev, and Xiong]{DBLP:conf/acl/YinRX21}
Wenpeng Yin, Dragomir~R. Radev, and Caiming Xiong.
\newblock Docnli: {A} large-scale dataset for document-level natural language inference.
\newblock In \emph{Findings of the Association for Computational Linguistics: {ACL/IJCNLP} 2021, Online Event, August 1-6, 2021}, 2021.
\newblock URL \url{https://doi.org/10.18653/v1/2021.findings-acl.435}.

\bibitem[Yu et~al.(2023)Yu, Li, Yao, Pang, Zhou, Xiao, Meng, and Zhou]{yu2023personality}
Mo~Yu, Jiangnan Li, Shunyu Yao, Wenjie Pang, Xiaochen Zhou, Zhou Xiao, Fandong Meng, and Jie Zhou.
\newblock Personality understanding of fictional characters during book reading.
\newblock \emph{arXiv preprint arXiv:2305.10156}, 2023.

\bibitem[Yu et~al.(2025{\natexlab{a}})Yu, Liu, Wu, Chung, Zhang, Li, Yeung, and Zhou]{yu2025stochastic}
Mo~Yu, Lemao Liu, Junjie Wu, Tsz~Ting Chung, Shunchi Zhang, Jiangnan Li, Dit-Yan Yeung, and Jie Zhou.
\newblock The stochastic parrot on llm's shoulder: A summative assessment of physical concept understanding.
\newblock \emph{arXiv preprint arXiv:2502.08946}, 2025{\natexlab{a}}.

\bibitem[Yu et~al.(2025{\natexlab{b}})Yu, Zhang, Qiao, Yin, Li, Wang, Chen, Zheng, Liang, and Sun]{yu2025sequential}
Yifei Yu, Qian-Wen Zhang, Lingfeng Qiao, Di~Yin, Fang Li, Jie Wang, Zengxi Chen, Suncong Zheng, Xiaolong Liang, and Xing Sun.
\newblock Sequential-niah: A needle-in-a-haystack benchmark for extracting sequential needles from long contexts.
\newblock \emph{arXiv preprint arXiv:2504.04713}, 2025{\natexlab{b}}.

\bibitem[Zhang et~al.(2024)Zhang, Chen, Hu, Xu, Chen, Hao, Han, Thai, Wang, Liu, and Sun]{DBLP:conf/acl/ZhangCHXCH0TW0024}
Xinrong Zhang, Yingfa Chen, Shengding Hu, Zihang Xu, Junhao Chen, Moo~Khai Hao, Xu~Han, Zhen~Leng Thai, Shuo Wang, Zhiyuan Liu, and Maosong Sun.
\newblock $\infty$bench: Extending long context evaluation beyond 100k tokens.
\newblock In \emph{Proceedings of the 62nd Annual Meeting of the Association for Computational Linguistics (Volume 1: Long Papers), {ACL} 2024, Bangkok, Thailand, August 11-16, 2024}, 2024.
\newblock URL \url{https://doi.org/10.18653/v1/2024.acl-long.814}.

\bibitem[Zhao et~al.(2024)Zhao, Andriushchenko, Croce, and Flammarion]{zhao2024long}
Hao Zhao, Maksym Andriushchenko, Francesco Croce, and Nicolas Flammarion.
\newblock Long is more for alignment: A simple but tough-to-beat baseline for instruction fine-tuning.
\newblock In \emph{International Conference on Machine Learning}, pp.\  60674--60703. PMLR, 2024.

\bibitem[Zhou et~al.(2025)Zhou, Wang, Yu, Yue, Lu, Li, Zhou, Zhang, Zhou, and Lam]{zhou2025essence}
Chulun Zhou, Qiujing Wang, Mo~Yu, Xiaoqian Yue, Rui Lu, Jiangnan Li, Yifan Zhou, Shunchi Zhang, Jie Zhou, and Wai Lam.
\newblock The essence of contextual understanding in theory of mind: A study on question answering with story characters.
\newblock \emph{arXiv preprint arXiv:2501.01705}, 2025.

\bibitem[Zhou et~al.(2023)Zhou, Liu, Xu, Iyer, Sun, Mao, Ma, Efrat, Yu, Yu, et~al.]{zhou2023lima}
Chunting Zhou, Pengfei Liu, Puxin Xu, Srinivasan Iyer, Jiao Sun, Yuning Mao, Xuezhe Ma, Avia Efrat, Ping Yu, Lili Yu, et~al.
\newblock Lima: Less is more for alignment.
\newblock \emph{Advances in Neural Information Processing Systems}, 36:\penalty0 55006--55021, 2023.

\end{thebibliography}
\bibliographystyle{iclr2025_conference}

\newpage
\appendix
\section{Assessment of Assessments: What Makes a Long-Context Benchmark Truly Meaningful?}
\label{app:criteria}

For each criterion discussed in Section~\ref{sec:intro}, we first go through the definition and examples of the criterion, then propose its delegate measure:

\paragraph{Beyond Memorization.}
\begin{wrapfigure}{r}{0.35\textwidth}
  \centering
  \vspace{-0.2in}
  \includegraphics[width=0.33\textwidth]{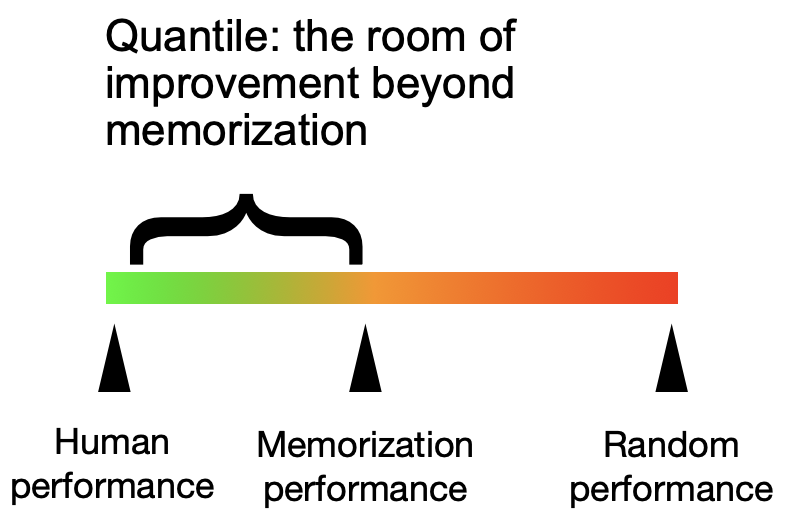}
  \caption{\small Illustration of the measurement for the criterion of \emph{Beyond Memorization}.}
  \label{fig:beyond_mem}
  \vspace{-0.255in}
\end{wrapfigure}
LLMs are known to memorize large amounts of training data~\citep{tirumala2022memorization,deletang2023language,ilya2023}. For popular texts (\emph{e.g.}, widely read books), models may recall content or associated analysis from pretraining, bypassing the need for actual comprehension. As training datasets continue to expand, this issue becomes increasingly problematic.
Therefore, a benchmark should be designed such that it cannot be solved purely through memorized knowledge, ensuring that the full length and structure of the context remain necessary for reasoning.

$\bullet$ \emph{Measurement:} 
We first evaluate the memorization performance of GPT-4o by having it answer questions without using RAG. We then compute its quantile within the range defined by human and random performance. 
This measurement is illustrated in Figure~\ref{fig:beyond_mem}.

\paragraph{Global Dependency.}
\begin{wrapfigure}{r}{0.35\textwidth}
  \centering
  \vspace{-0.2in}
\includegraphics[width=0.33\textwidth]{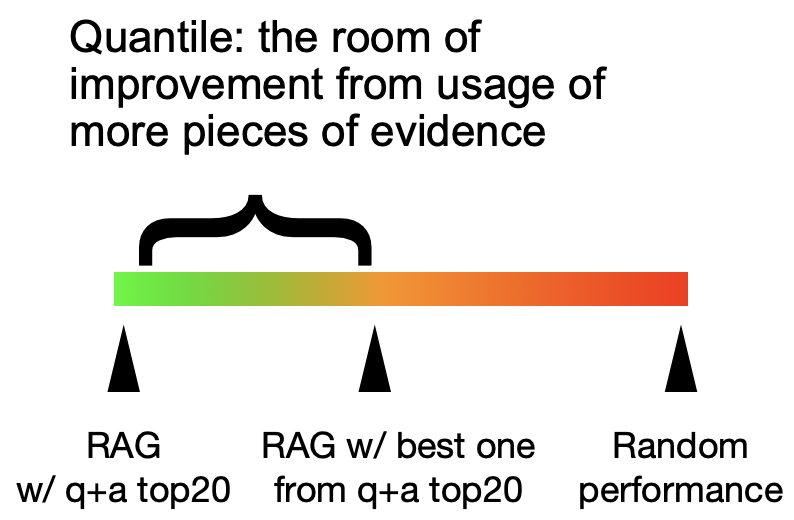}
  \caption{\small Illustration of the measurement for the criterion of \emph{Global Dependency}.}
  \label{fig:global_dep}
  \vspace{-0.25in}
\end{wrapfigure}
The task should require aggregating information from multiple pieces of evidence that are scattered across the context or exhibit global dependencies. Otherwise, it reduces to a short-context problem solvable by retrieving the relevant passage for a given question. In such cases, the task becomes more about improving retrieval quality than long-context understanding ability.

$\bullet$ \emph{Measurement:}  
This dimension differs slightly from the previous one in how the performance interval is defined. To evaluate how much better a model performs compared to using a single document, the reference point should be the model’s performance with sufficient evidence, rather than human performance. To approximate this, we retrieve the top-20 documents using both the question and the answer, and treat this as the upper bound of performance with sufficient evidence.
We then ask GPT-4o to select the best supporting document from the top-20 retrieved using both the question and answer, treating it as the \textbf{strongest single piece of evidence}. We compute the quantile of the model's RAG performance using only this selected document, relative to the interval defined by the QA-top-20 RAG performance and the random baseline.

\paragraph{Depth of Reasoning.}
\begin{wrapfigure}{r}{0.35\textwidth}
  \centering
  \vspace{-0.2in}
\includegraphics[width=0.33\textwidth]{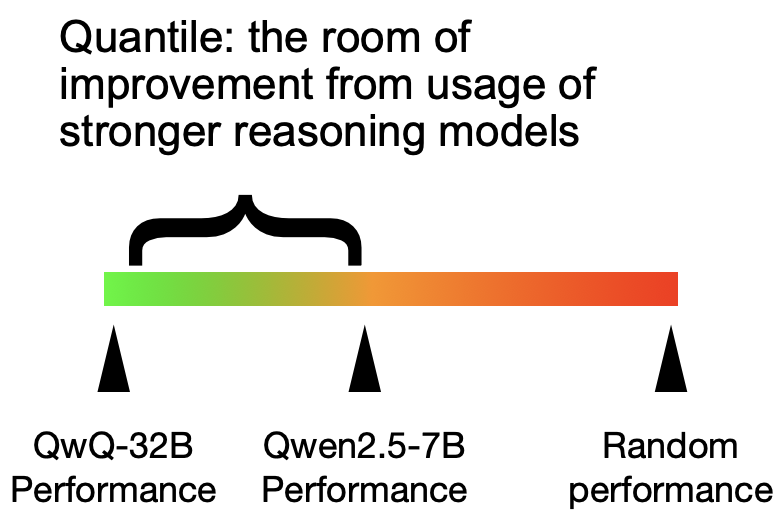}

  \caption{\small Illustration of the measurement for the criterion of \emph{Depth of Reasoning}.}
  \label{fig:reasoning_depth}
  \vspace{-0.2in}
\end{wrapfigure}

By nature, long-context reasoning involves synthesizing multiple pieces of evidence across the input. The complexity of reasoning—especially multi-step deduction—is closely tied to task difficulty.
Many existing benchmarks emphasize shallow reasoning, primarily requiring decomposition or enumeration (e.g., counting mentions of a fact~\citep{xu2025llm}, or multi-hop QA where the questions are often constructed as sequences of relatively simple sub-questions~\citep{yang2018hotpotqa}.
This limits the need for global reasoning and makes tasks easier than intended.

$\bullet$ \emph{Measurement:}  
Similar to the previous dimension, the key in this measurement lies in identifying a representative performance interval and a suitable reference model for computing the quantile. We choose QwQ-32B as a strong reasoning model and treat its RAG performance using the top-20 documents retrieved with both the question and answer (as defined in the previous measurement) as the upper bound. We then compute the quantile of Qwen2.5-7B under the same RAG setting. This gap reflects the potential for improvement attributable to stronger reasoning capabilities. Thus, the deeper the reasoning required by a dataset, the larger this gap is expected to be.

Note that this proxy becomes less informative if the dataset is so challenging that neither model significantly outperforms the random baseline. However, as shown in Table~\ref{tab:benchmark_comparison}, this is not currently the case. Therefore, this measurement remains a meaningful indicator.

\paragraph{Beyond Summarization/Salience.}
This often-overlooked criterion is crucial: Tasks that can be resolved simply by generating a summary of a long input are less challenging and may no longer probe deep understanding, especially given recent advances in summarization (\emph{e.g.}, hierarchical or iterative methods~\citep{chang2023booookscore}).
In such cases, the challenge of long-context understanding is effectively reduced to a summarization task over short segments, creating a shortcut.
A high-quality benchmark should instead require attention to fine-grained details that go beyond high-level abstraction, or it can be reduced to a relatively easier task of summarization thus is less meaningful.

$\bullet$ \emph{Measurement:}  
This dimension is measured by comparing the performance of RAG using original text chunks against RAG using chunk summaries of the same input length.
We do not use quantiles here but instead directly report the relative improvement, as some datasets actually perform worse when using the original texts, indicating that the task questions primarily target salient events, making summarization sufficient for answering.

\paragraph{Human-Machine Gap.}
\begin{wrapfigure}{r}{0.35\textwidth}
  \centering
  \vspace{-0.2in}
\includegraphics[width=0.33\textwidth]{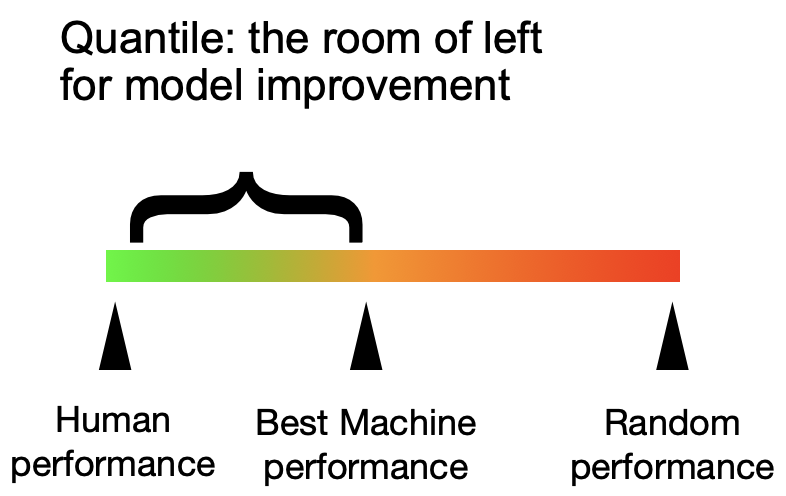}
  \caption{\small Illustration of the measurement for the criterion of \emph{Human-Machine Gap}.}
  \label{fig:human_machine_gap}
\end{wrapfigure}
To highlight essential capabilities that general-purpose intelligent systems ought to possess,
a meaningful benchmark should show a significant gap between humans and machines. At the same time, poor human accuracy or inter-annotator agreement typically signals low annotation quality or high subjectivity — both detrimental to robust evaluation.

$\bullet$ \emph{Measurement:}
We identify the best machine performance from the evaluations in the \emph{Beyond Memorization} and \emph{Beyond Summarization} dimensions and compute its quantile within the interval defined by human performance and the random baseline.

\begin{table*}
\small
\centering
\resizebox{1.\columnwidth}{!}{
\begin{tabular}{@{}l|l|ccccc}
\toprule
\multirow{2}{*}{\bf Criterion} & \multirow{2}{*}{\bf Metric} & \multicolumn{5}{c}{\bf Datasets} \\
&&\bf InfBench.MC &\bf DetectiveQA & \bf NarrativeQA&\bf NoCha (public)& \bf \datasetname\\
\midrule
-- & Random & 25.0 &25.0 & 7.1 & 50.0 & 50.0 \\
-- & Human Performance & 100.0$^*$ & 94.0& 64.3 & 97.0 & 82.0\\
-- & Human-Random Gap & 75.0 & 69.0& 57.2 & 47.0 & 32.0\\
\midrule
\multirow{2}{*}{Beyond Memorization} & Memorization Performance & 77 & 50.5 & 36.1 & 80.2 & 56 \\
&\it Quantile of Memorization & 30.7 &  63.1 & 49.3 & 35.8 & \bf 81.3\\
\midrule
\multirow{3}{*}{Global Dependency} & RAG (qa-top20)& 85.0 & 73.3  & 35.9 & 73.8 & 59.0\\
& RAG (best one from qa-top20) & 84.0 &  55.2 & 32.9 & 57.4 & 53.0\\
&\it Quantile of RAG (best-one) & 1.7 & 37.4 & 10.6 & 68.9 & \bf 66.7\\
\midrule
\multirow{3}{*}{Depth of Reasoning} & Qwen2.5-7B w/ RAG qa-top20 & 68.0 & 68.6 & 24.0 & 60.3 & 52.0 \\
& QwQ-32B w/ RAG qa-top20  & 86.0 & 81.0 & 42.3 & 66.7& 56.0\\
&\it Quantile of Qwen2.5-7B & 29.5 & 22.1 & 52.1 & 38.1 & \bf 66.7 \\
\midrule
\multirow{3}{*}{Beyond Summarization} & RAG over chunks (max-10K) & 83.0 & 77.4 & 23.9 & 68.3 & 61.0\\
& RAG over chunk summaries (max-10K) & 80.1 & 77.2 & 27.1 &73.8 & 59.0\\
&\it Relative Improvement &\bf 3.6 & 0.3 & -11.8 & -7.5 & 3.4 \\
\midrule
\multirow{2}{*}{Human-Machine Gap} & Best Machine Performance & 83.0 & 77.4 & 36.1 & 80.2 & 61.0\\
&\it Quantile of Machine Performance & 22.7 & 24.0 &49.3 &35.8 &\bf 65.6 \\
\midrule
Average Doc Length & -- & 228K & 97K & 104K & 153K & \bf 408K \\
\bottomrule
\end{tabular}
}
\caption{\small{The detailed experimental results we used to compute the measurement of representative criteria for good long-context understanding benchmarks. Except for the Qwen2.5-7B and QwQ-32B results, we generate the results with Qwen3-Embedding-8B and GPT4o. *The paper did not provide the human performance but claimed humans can achieve near-perfectly so we trust their claim.}}
\label{tab:benchmark_comparison}
\end{table*}

\paragraph{Compared Datasets.}
We compare our \datasetname with representative benchmarks from prior work, including~\citep{kovcisky2018narrativeqa,DBLP:conf/emnlp/KarpinskaTLGI24,zhao2024long,xu2025detectiveqa}. These widely used datasets span multiple genres such as classic and detective novels, support both English and Chinese languages, and cover a range of task formats including free-form QA, multiple choice, and true-or-false questions. For a fair comparison, we sample $\sim$100 questions from each benchmark: InfBench.MC (25 books), DetectiveQA (10 books), and NarrativeQA (10 books from the development set), ensuring a similar number of examples across datasets.
\section{Details of the Dataset}
\label{app:dataset}

In this section we provide the detailed characters and questions used in our study in both English and Chinese dataset. The public subset will be included in our code and data released upon publication. 

\subsection{Prompt for Generating Prequels}
\label{app:prequel_generation}
We simple prompt the LLMs with no additional context to generate the prequels for human annotation.
Figure~\ref{fig:prequel_gen_prompt_template} demonstrates the English translation of our prompt.

\begin{figure}[ht]
    \centering
\begin{lstlisting}[language=prompt, basicstyle=\ttfamily\tiny]
You are a writing assistant who is deeply familiar with the plots of various renowned literary works and highly skilled in literary analysis.

Please write a prequel story for the character "{{ [[[char]]] }}" based on the plot of the novel {{ [[[book_name]]] }}. The story should provide a background that does not contradict the original work. It should explain all of the character’s major actions in the original novel, their key personality traits, and their major relationships. Your prequel should focus on reasonably answering unresolved questions about the character that are raised in the original novel. These include: important issues that cannot be explained using the original content alone, plot points that appear inconsistent with the character’s established traits or internal logic, and significant past experiences that are mentioned but not elaborated upon.

The background story must remain coherent with the original plot and should not introduce settings or elements that are clearly inconsistent with the style of the original work.

You only need to write an outline. The outline should summarize key elements of the character’s backstory and major life events. Each item in the outline should include specific story-level details, rather than general concepts or attributes. The total length of the outline should be within 1,000 words. Please present your answer in a Markdown list format.
\end{lstlisting}
    \caption{The prompt we used to generate prequels for human annotation (English translation).}
    \label{fig:prequel_gen_prompt_template}
\end{figure}

\subsection{Selected Characters}
\label{app:dataset_char}

Table~\ref{tab:characters} lists the main characters used to in our \datasetname for annotation.

\begin{table*}[ht!]
\small
\centering
\resizebox{1.\columnwidth}{!}{
\begin{tabular}{ll||p{0.6\linewidth}} 
\toprule
\bf Name& \bf Author & \bf Characters\\
\midrule
The Count of Monte Cristo & Alexandre Dumas & \emph{Faria, Noirtier, General de Quesnel}\\
Demi-Gods and Semi-Devils & Louis Cha & \emph{Bo Murong, Yellow-Browed Monk, Sweeper Monk} \\
The Return of the Condor Heroes & Louis Cha & \emph{Yin Zhiping, Yelv Qi, Huo-Du, Gongsun Zhi} \\
Investiture of the Gods & Xu Zhonglin & \emph{Wen Zhong, Zhao Gongming, Zhang Kui, Huang Feihu} \\
Romance of the Three Kindoms & Luo Guanzhong & \emph{Xiahou Yuan, Taishi Ci} \\
Love in the Time of Cholera & Gabriel García Márquez & \emph{Don Pius V Loayza, Lotario Thugut, Rosalba} \\
Pinball, 1973 & Haruki Murakami & \emph{The Rat, J, Naoko} \\ 
Rebecca & Daphne du Maurier & \emph{Mrs. Van Hopper, Mrs. Danvers, Jack Favell} \\ 
In Search of the Castaways & Jules Verne & \emph{Jacques Paganel, Tom Ayrton/Ben Joyce, Thalcave, Kai-Koumou}
\\
The Redeemer & Jo Nesbo & \emph{Christo Stankic, Martine Eckhoff, Jon Karlsen} \\
Drawing Sword & Du Liang & \emph{Tian Moxuan, Ding Wei, Yamamoto Kazuki} \\
Dwelling Narrowness & Liu Liu & \emph{Chen Sifu, Mrs. Song, Su Chun} \\
Distant Saviour & Dou Dou & \emph{Liu Bing, Han Chufeng, Xiao Yawen, Lin Yufeng} \\
Attack on Titan & Hajime Isayama \& the Fans & \emph{Eren Kruger} \\

\bottomrule
\end{tabular}
}
\caption{\small{Main characters considered in our \datasetnamens.}}
\label{tab:characters}
\end{table*}

\section{Implementation Details}
\label{app:implementation_details}
\subsection{Hyperparameters of In-Domain training}
We apply LoRA~\citep{hu2021lora} to all available target modules for in-domain training with RAG and non-RAG settings. The training is conducted on the dataset excluding the human-study subset, resulting in a total of 695 instances. Since the average token length in the RAG setting is above 20k, we perform LoRA tuning on both settings using 360-LLaMA-Factory~\citep{360-llama-factory}. We fine-tune each model for 2 epochs with a batch size of 1 on a single machine with 8 H20 GPUs. The dimension of LoRA's attention layer is set to 64, while $\alpha$ is configured to 128. In addition, the learning rate and warmup ratio are set to 1e-4 and 0.1 respectively.

\subsection{Prompt Templates}
We use the prompt template in Figure \ref{fig:cn_rag_prompt_template} and \ref{fig:en_rag_prompt_template} for the RAG experiment, depending on whether the instance originates from a Chinese book or an English book. For the ``No Context'' experiment, we remove the original excerpt part both in the ICL examples and the actual instance.
The exact content of ICL in both languages is in Figure \ref{fig:cn_egs} and \ref{fig:en_egs}.

\section{Additional Results}
\label{app:additional_results}

\subsection{Illustrative Examples for Qualitative Study}
\label{app:qualitative}
\begin{itemize}[noitemsep,nolistsep,leftmargin=*]
\item \emph{Subjective Example -- Fuzzy Logic}: \emph{Example~\ref{ex:fuzzy_logic}} shows an example that our annotators felt difficult to arrive an agreement due to the fuzzy reasoning. This setting results in the fact that the character appears in the story at around 120 years old. The novel does feature several characters in their 100s who remain healthy and skilled in combat. Still, as the most advanced martial artist in the story, whether 120 is a reasonable age is a matter of fuzzy logic.
\vspace{0.1in}
\begin{exampletable}[ex:fuzzy_logic]{{\small An example of disagreement among human annotators due to fuzzy reasoning.}}
\resizebox{1.\columnwidth}{!}{
\begin{tabular}{p{18cm}}
\hline
\textbf{Book:} Demi-Gods and Semi-Devils \quad \textbf{Char:} Sweeper Monk \quad \textbf{Groundtruth:} Consistent \newline \textbf{Prequel:} His birth name was Li Xuankong, a collateral member of the imperial family of Later Zhou during the late Five Dynasties period. As a child, he was orphaned by war and taken in by a hermit from Mount Song, who was secretly a first-generation disciple of the Xiaoyao Sect (unaffiliated with the internal conflict within Wuya Zi’s lineage).\\
\hline
\end{tabular}
}
\end{exampletable}
\vspace{0.2in}
\item \emph{Subjective Example -- Different Interpretation of Persona (Example~\ref{ex:subjective_interpretation})}: In the story, the Sweeper Monk is not a formal disciple of Shaolin; he is more like a hermit.
So the disagreement arises that when others mention that many Shaolin disciples had learned Xiaowuxiang Gong, whether the Sweeper Monk would interpret that as referring to himself? The annotators would project themselves into the character’s perspective when making predictions, which introduces subjectivity.
\vspace{-0.02in}
\begin{exampletable}[ex:subjective_interpretation]{{\small An example of disagreement among human annotators due to subjective interpretation of character persona.}}
\resizebox{1.\columnwidth}{!}{
\begin{tabular}{p{18cm}}
\hline
\textbf{Book:} Demi-Gods and Semi-Devils \quad \textbf{Char:} Sweeper Monk \quad \textbf{Groundtruth:} Consistent \newline \textbf{Prequel:} As a young prodigy, he simultaneously studied the incomplete version of Shaolin's Yijin Jing and the Xiaoyao Sect's Xiaowuxiang Gong, though he never formally joined any sect.\\
\hline
\end{tabular}
}
\end{exampletable}
\vspace{0.2in}
\item \emph{Example that RAG Performs Worse}: \emph{Example~\ref{ex:rag_failure}} gives an example in which the vanilla LLM can predict correct answer but fails when equipped with RAG. The prequel is consistent because it does not violate any part of the story. Specifically, in a flashback scene from the original novel, it is revealed that Murong Bo’s mother subjected him to harsh training in order to raise him as a future monarch for national restoration.
When the RAG system fails to recognize that this passage describes Murong Bo’s childhood, it tends to rely solely on the more prominent information in the input (e.g., Document 0) to make its judgment.
\vspace{0.1in}
\begin{exampletable}[ex:rag_failure]{{\small An example that RAG makes the performance worse.}}
\resizebox{1.\columnwidth}{!}{
\begin{tabular}{p{18cm}}
\hline
\textbf{Book:} Demi-Gods and Semi-Devils \quad \textbf{Char:} Murong Bo \quad \textbf{Groundtruth:} Consistent \newline \textbf{Prequel:} On her deathbed, his mother pressed a bloodstained jade pendant engraved with the Yan dynasty dragon emblem into his hand, repeatedly stressing that “the Xianbei Murong clan and the Zhao-Song regime cannot coexist.”\newline \textbf{Model's Reason for Contradict:} In the original novel, Murong Bo’s restorationist education clearly stems from a paternal lineage. His mother is portrayed merely as an ordinary family member.
The added element of "the mother reinforcing the restorationist ideology" has no textual basis and contradicts the Murong clan’s patrilineal transmission system of “from father to son, from grandfather to grandson” (see Document 0).\\
\hline
\end{tabular}
}
\end{exampletable}
\vspace{0.2in}
\item \emph{Example with Correct Answer but Incorrect Reasoning}: In \emph{Example~\ref{ex:incorrect_reason}}, the model correctly answers the question, but bases on incorrect reasons.
The actual basis for judging this as a contradiction lies in the fact that both characters are portrayed as emphasizing personal martial prowess, with no demonstration of strategic ability or any depiction of possessing military treatises in the original novel. The act of gifting The Six Secret Teachings (Liu Tao)—a text regarded in Chinese tradition as one of the most profound and authoritative works on military strategy—stands in sharp contrast to their characterization.

\quad The model's explanation is invalid because it fails to grasp that Wu Anguo was representing Beihai's forces (under Kong Rong) in the battle at Hulao Pass. Instead, it superficially interprets the text and incorrectly assumes that Wu Anguo's sphere of activity was limited to Hulao Pass.

\quad This reflects the same underlying issue as in the RAG failure example: LLMs are heavily influenced by the provided context and tend to overanalyze non-informative clues, often becoming overly literal and fixating on minor wording differences.
\end{itemize}

\begin{exampletable}[ex:incorrect_reason]{{\small An example that LLMs make correct prediction with incorrect reasoning.}}
\resizebox{1.\columnwidth}{!}{
\begin{tabular}{p{18cm}}
\hline
\textbf{Book:} Romance of the Three Kingdoms \quad \textbf{Char:} Taishi Ci  \quad \textbf{Groundtruth:} Contradict \newline \textbf{Prequel:} At the age of seventeen, while escorting his mother on a visit to relatives, he encountered remnants of the Yellow Turban rebels. In the midst of a fierce battle, he rescued Wu Anguo, the gate captain of Beihai, who was under siege. Moved by his bravery, Wu Anguo gifted him a handwritten copy of The Six Secret Teachings by Jiang Ziya.\newline \textbf{Model's Reason for Contradict:} In the original novel, 
Wu Anguo’s only appearance in Romance of the Three Kingdoms is in Chapter 5, where he has his wrist severed by Lü Bu during the battle at Hulao Pass. He is never mentioned again, nor is there any record of him serving as a gate captain in Beihai.
This character setting not only fabricates an encounter between Wu Anguo and a teenage Taishi Ci (violating Rule 1), but also forcibly relocates Wu Anguo—a defeated general from the Hulao Pass battle—to a post in Beihai, disrupting the narrative continuity of that battle (violating Rule 2).\\
\hline
\end{tabular}
}
\end{exampletable}

\subsection{Performance Decomposition to Books}
\label{app:decomposition}

\begin{table*}[h!]
\small
\centering
\resizebox{1.\columnwidth}{!}{
\begin{tabular}{l||c|cccccc} 
\toprule
\bf Book Name& \bf \#Samples & \bf GPT4o & \bf + RAG & \bf R1 & \bf + RAG & \bf Gemini-2.5-Pro & \bf + RAG\\
\midrule
The Count of Monte Cristo & 54 & 66.7& 70.4& 68.5& 70.4& 72.2& 75.9\\
Demi-Gods and Semi-Devils & 78 & 50.0 & 65.4 & 56.4 & 66.7 & 75.6 & 64.1\\
The Return of the Condor Heroes & 44 & 56.8 & 61.4 & 65.9 & 65.9 & 72.7 & 77.3\\
Investiture of the Gods & 94 & 59.6 & 62.8 & 61.7 & 67.0 & 69.1 & 69.1\\
Romance of the Three Kingdoms & 28 & 64.3 & 46.4 & 64.3 & 75.0 & 57.1 & 46.4\\
Love in the Time of Cholera & 15 & 46.7& 53.3& 66.7& 60.0& 73.3& 53.3\\
Pinball, 1973 & 22 & 68.2 & 68.2 & 54.5 & 68.2 & 63.6 & 54.5\\
Rebecca & 82 & 67.1& 64.6& 68.3& 47.6& 64.6& 65.9\\
In Search of the Castaways & 86 & 70.9& 62.8& 74.4& 48.8& 51.2& 47.7\\
The Redeemer & 82 & 42.7& 53.7& 48.8& 58.5& 64.6& 61.0\\
Drawing Sword & 70 & 64.3 & 65.7 & 64.3 & 60.0 & 67.1 & 62.9\\
Dwelling Narrowness & 92 & 70.7 & 60.9 & 65.2 & 57.6 & 64.1 & 53.3\\
Distant Saviour & 48 & 54.2 & 60.4 & 56.2 & 64.6 & 60.4 & 66.7\\
\bottomrule
\end{tabular}
}
\caption{\small{Performance decomposition to books. We report accuracy on each book.}}
\label{tab:decomposed_results}
\end{table*}

\begin{figure}[ht]
    \centering
\begin{CJK}{UTF8}{gbsn}
\begin{lstlisting}[language=prompt, basicstyle=\ttfamily\tiny]
%*你是一位对各种著名文学作品剧情了如指掌、且精通文学分析的写作助手。*)

%*我正在为小说*)“{{ [[[book_name]]] }}”%*中的人物*)“{{ [[[char]]] }}”%*写该人物的前传故事，你将阅读到我写的前传中的一条剧情设定，并帮助我判断这一项剧情是否会与原书中的剧情和设定形成矛盾。*)

%*判断规则如下：*)
1. %*如果原书中存在一个细节与我提供的剧情设定直接构成矛盾，请回答“矛盾”；*)
2. %*如果我提供的剧情设定导致原书中部分剧情或者人物行为动机整体不合理，请回答“矛盾”；*)
3. %*如果我提供的剧情设定与原书中剧情不构成矛盾，但是与原书的设定风格不统一，请回答“矛盾”；*)
4. %*如果我提供的剧情设定与原书不存在以上类型的明显矛盾，请回答“一致”；*)
5. %*请在你的回答中给出你的判断，并提供简要的判断依据；*)
6. %*你将被提供一些原书中可能的相关片段作为参考，但是这些内容可能不够全面，因此请结合你对该书的知识和所提供的原文片段进行回答；*)
7. %*你应该只基于小说原书的内容进行判断，不要考虑其他相关作品、改编或者人物历史原型。*)
8. %*在判断时，你需要假设整个故事由该剧情设定的前传故事和原书直接承接构成，两者中间没有其他故事发生。*)
9. %*除非某个人物的台词在后面被显式证明是谎言，否则你应该将其认为是事实，同理所有旁白也应被认为是事实。*)

%*以下为该任务的一些示例（每个示例对应的相关原文较长，故略去）：*)
<example_1>
<<<[example_1]>>>
</example_1>

<example_2>
<<<[example_2]>>>
</example_2>

<example_3>
<<<[example_3]>>>
</example_3>

<example_4>
<<<[example_4]>>>
</example_4>

<example_5>
<<<[example_5]>>>
</example_5>

%*请根据以上示例，完成以下任务：*)
[%*前传剧情设定*)]
%*小说：*){{ [[[book_name]]] }}
%*人物：*){{ [[[char]]] }}
%*相关原文：*)
<chunks>{{ [[[chunks]]] }}</chunks>
%*标题：*){{ [[[subject]]] }}
%*设定：*){{ [[[content]]] }}

%*请仿照以上例子的输出格式，给出你的回答。*)
\end{lstlisting}
\end{CJK}
    \caption{Chinese prompt template used for the ``RAG'' experiment in Table \ref{tab:human_study_set_results}, with contents of {\textcolor{brown}{\small\textbf{\texttt{[example\_i]}}}} presented in Figure \ref{fig:cn_egs}.}
    \label{fig:cn_rag_prompt_template}
\end{figure}

\begin{figure}[ht]
    \centering
    \begin{CJK}{UTF8}{gbsn}
    \begin{lstlisting}[language=prompt, basicstyle=\ttfamily\tiny]
<<<[example_1]:>>>
%*[前传剧情设定]*)
%*小说：西游记*)
%*人物：金角大王与银角大王*)
%*相关原文：（省略）*)
%*标题：下界因果*)
%*设定：五百年前孙悟空踢翻八卦炉时遗落三块火砖（对应平顶山三峰），受老君密令下界回收火砖*)

%*[输出]*)
Think: %*原书中八卦炉遗落火砖形成的是火焰山，而不是平顶山。因此符合规则第一条，与原书细节矛盾。*)
Answer: %*矛盾*)


<<<[example_2]:>>>
%*[前传剧情设定]*)
%*小说：西游记*)
%*人物：黄狮精*)
%*相关原文：（省略）*)
%*标题：初生与灵根渊源*)
%*设定：机缘巧合下，它得获一只“黄玉狮爪”法器雏形（后被其修炼为主要武器），为日后战斗增添独特威力。*)

%*[输出]*)
Think: %*原书中黄狮精从来没有用过黄玉狮爪这种法器，他的武器是四明铲。因此该设定符合规则第二条，与原书多处剧情构成整体不一致。*)
Answer: %*矛盾*)


<<<[example_3]:>>>
%*[前传剧情设定]*)
%*小说：三国演义*)
%*人物：夏侯渊*)
%*相关原文：（省略）*)
%*标题：寒门崛起*)
%*设定：得异人传授神行法道术，因此日后带兵可以“三日五百，六日一千”*)

%*[输出]*)
Think: %*虽然与原书剧情不矛盾，但是原书（三国演义）整体保持写实风格，其中的将军带兵作战主要凭借个人武勇，从来没有使用法术作战的例子。因此该设定符合规则第三条，与原书的设定风格不统一。*)
Answer: %*矛盾*)


<<<[example_4]:>>>
%*[前传剧情设定]*)
%*小说：三国演义*)
%*人物：徐庶*)
%*相关原文：（省略）*)
%*标题：初露锋芒*)
%*设定：中平元年（184年）黄巾之乱，护送母亲逃亡途中目睹官军屠村，首次杀人（17岁）。因救被掳妇女得罪当地豪强，开始化名"单福"。*)

%*[输出]*)
Think: %*原书中有徐庶年少任侠杀人，和化名单福的描写，但是没有给出具体细节。该设定补全了原书缺少的细节，且没有构成矛盾。因此是一个与原著存在高度兼容性的剧情设定。*)
Answer: %*一致*)


<<<[example_5]:>>>
%*[前传剧情设定]*)
%*小说：西游记*)
%*人物：黄狮精*)
%*相关原文：（省略）*)
%*标题：短暂拜师与炼法*)
%*设定：为求更高法力，黄狮精曾私自闯入周边其他山头，与多位小妖对决，锤炼法术，也树下冤仇。*)

%*[输出]*)
Think: %*原文中在相关章节从未提过黄狮精与周边其他山头的冤仇。该设定与原书剧情脱离，但也因此不会与原书剧情构成矛盾。因此是一个与原著剧情无紧密联系的剧情设定。*)
Answer: %*一致*)
    \end{lstlisting}
    \end{CJK}
    \caption{Content of the in-context chinese examples in Figure \ref{fig:cn_rag_prompt_template}.} 
    \label{fig:cn_egs}
\end{figure}

\begin{figure}[ht]
    \centering
\begin{lstlisting}[language=prompt, basicstyle=\ttfamily\tiny]
You are a writing assistant who is deeply familiar with the plots of various renowned literary works and highly skilled in literary analysis.

I am writing a prequel story for the character “{{ [[[char]]] }}” from the novel “{{ [[[book_name]]] }}”. You will read a plot point from this prequel and help me determine whether it conflicts with the original novel’s plot and setting.

Please follow the rules below when making your judgment:

1. If a detail in the original novel directly contradicts the plot point I provide, please respond with “Contradict”;
2. If the plot point I provide causes part of the original plot or the character’s motivations to become broadly unreasonable, please respond with “Contradict”;
3. If the plot point does not contradict the original plot but is inconsistent with the tone or stylistic setting of the novel, please respond with “Contradict”;
4. If none of the above conflicts exist between the plot point and the original novel, please respond with “Consistent”;
5. In your response, clearly state your judgment and briefly explain your reasoning;
6. You will be provided some snippets from the original book for reference, which may not be comprehensive. Please consider both the snippets and your knowledge about the book to provide the answer;
7. You should base your judgment only on the content of the original novel — do not consider adaptations, derivative works, fandoms, or historical inspirations behind the characters;
8. You should assume that the prequel is followed immediately by the original story, with no additional text or events in between;
9. Unless a character's statements are later explicitly revealed to be deceptive, they should be treated as factual, similar for narrative expositions.

Here are some examples for reference (the corresponding original texts are long, so they are omitted):
<example_1>
<<<[example_1]>>>
</example_1>

<example_2>
<<<[example_2]>>>
</example_2>

<example_3>
<<<[example_3]>>>
</example_3>

<example_4>
<<<[example_4]>>>
</example_4>

<example_5>
<<<[example_5]>>>
</example_5>

Please follow the format above and complete the task below:
[Prequel Plot Point]
Novel: {{ [[[book_name]]] }}
Character: {{ [[[char]]] }}
Original Excerpt:
<chunks>{{ [[[chunks]]] }}</chunks>
Title: {{ [[[subject]]] }}
Setup: {{ [[[content]]] }}

Please give your answer following the format shown in the examples above.
\end{lstlisting}
    \caption{English prompt template used for the ``RAG'' experiment in Table \ref{tab:human_study_set_results}, with contents of {\textcolor{brown}{\small\textbf{\texttt{[example\_i]}}}}  presented in Figure \ref{fig:en_egs}.}
    \label{fig:en_rag_prompt_template}
\end{figure}

\begin{figure}[ht]
    \centering
\begin{lstlisting}[language=prompt, basicstyle=\ttfamily\tiny]
<<<[example_1]:>>>
[Prequel Plot Point]
Novel: The Chronicles of Narnia
Character: Eustace Scrubb
Original Excerpt: (omitted)
Title: Early Life
Setup: Born in London and raised as an only child, Eustace was taught to call his parents by their names —"Harold" and "Alberta"—rather than "Mother" and "Father".

[Output]
Think: In the original novel, the Scrubb family lives in Cambridge and Eustace was born in Cambridge England. Therefore, this detail is consistent with Rule 1 and contradicts a factual detail from the original story.
Answer: Contradict


<<<[example_2]:>>>
[Prequel Plot Point]
Novel: The Chronicles of Narnia
Character: Eustace Scrubb
Original Excerpt: (omitted)
Title: Adventure in Underworld
Setup: The Lady of the Green Kirtle directs Eustace to the Underland, helping him find Prince Rilian and kill the serpent responsible for the prince’s enslavement and intent on killing him. 

[Output]
Think: In the story, Eustace is actually directed by her to the giant’s castle and accidentally discovers the Underland. Since the Lady of the Green Kirtle is herself the serpent, it would be impossible for her to intentionally lead them to the Underland and help them find the prince. Therefore, this scenario matches Rule 2 and introduces inconsistencies across multiple parts of the original plot.
Answer: Contradict


<<<[example_3]:>>>
[Prequel Plot Point]
Novel: Romance of the Three Kingdoms
Character: Xiahou Yuan
Original Excerpt: (omitted)
Title: Rise from Humble Origins
Setup: He was taught magical speed-enhancing techniques by a mysterious hermit, which allowed him to march “500 li in three days, and 1000 li in six days.”

[Output]
Think: Although not contradictory to the plot, the original novel maintains a realistic tone—generals fight with martial skills and do not use magic. This setup violates Rule 3 by being stylistically inconsistent with the novel.
Answer: Contradict


<<<[example_4]:>>>
[Prequel Plot Point]
Novel: Harry Potter
Character: Sirius Black
Original Excerpt: (omitted)
Title: Belief and Value
Setup: Rejecting pure-blood elitism, Sirius frequently showed admiration for Gryffindor and spent time with Muggles and so-called "blood traitors" to annoy his parents even before attending Hogwarts.

[Output]
Think: In the original novel, Sirius is introduced as a key figure against pure-blood elitism and is sorted into Gryffindor in 1971. However, the relevant chapters do not describe these specific actions. This addition fills in missing details without introducing contradictions. Therefore, it is compatible with the original.
Answer: Consistent


<<<[example_5]:>>>
[Prequel Plot Point]
Novel: Journey to the West
Character: Yellow Lion Demon
Original Excerpt: (omitted)
Title: Brief Apprenticeship and Spell Training
Setup: Seeking greater power, he once secretly ventured to other nearby mountains, fought with lesser demons, honed his magic, and made enemies along the way.

[Output]
Think: The relevant chapters in the original text do not mention these rivalries. While this setup is detached from the original plot, it doesn’t contradict it. Thus, it is a side story that remains compatible with the canon.
Answer: Consistent
\end{lstlisting}
    \caption{Content of the in-context english examples in Figure \ref{fig:en_rag_prompt_template}.} 
    \label{fig:en_egs}
\end{figure}

\label{app:baseline_prompts}

\end{document}